\documentclass[11pt]{article}

\usepackage{acl}

\usepackage{times}
\usepackage{latexsym}
\usepackage{url}
\usepackage[T1]{fontenc}
\usepackage[utf8]{inputenc}
\usepackage{microtype}
\usepackage{xcolor}
\usepackage{graphicx}
\usepackage{booktabs}
\usepackage{tabularx}
\usepackage{multirow}
\usepackage{multicol}
\usepackage{amsmath}
\usepackage{amssymb}
\usepackage{utfsym}
\usepackage{comment}
\usepackage{gb4e}
\usepackage{hyperref}
\usepackage{cleveref}
\usepackage{subcaption}
\usepackage{soul}
\usepackage{array}
\usepackage{float}

\noautomath

\title{Humans and LLMs Diverge on Probabilistic Inferences}

\author{
  Gaurav Kamath$^{\alpha,\beta}$ \quad
  Sreenath Madathil$^{\alpha,\beta}$ \quad
  Sebastian Schuster$^{\gamma}$ \\
  \textbf{Marie-Catherine de Marneffe}$^{\tau}$ \quad
  \textbf{Siva Reddy}$^{\alpha,\beta, \delta}$ \\[1em]
  $^{\alpha}$McGill University \quad
  $^{\beta}$Mila -- Quebec AI Institute \\
  $^{\gamma}$University of Vienna \quad
  $^{\tau}$FNRS -- UCLouvain \quad
  $^{\delta}$Canada CIFAR AI Chair
}

\date{}

\begin{document}
\maketitle

\begin{abstract}
Human reasoning often involves working over limited information to arrive at probabilistic conclusions.
In its simplest form, this involves making an inference that is not strictly entailed by a premise, but rather only likely given the premise.
While reasoning LLMs have demonstrated strong performance on logical and mathematical tasks, their behavior on such open-ended, non-deterministic inferences remains largely unexplored. 
We introduce \textsc{ProbCopa}, a dataset of 210 handcrafted probabilistic inferences in English, each annotated for inference likelihood by 25--30 human participants. 
We find that human responses are graded and varied, revealing probabilistic judgments of the inferences in our dataset.
Comparing these judgments with responses from eight state-of-the-art reasoning LLMs, we show that models consistently fail to produce human-like distributions.
Finally, analyzing LLM reasoning chains, we find evidence of a common reasoning pattern used to evaluate such inferences.
Our findings reveal persistent differences between humans and LLMs, and underscore the need to evaluate reasoning beyond deterministic settings.\footnote{All data and code can be found at \url{github.com/McGill-NLP/probabilistic-reasoning}}
\end{abstract}

\begin{figure}
    \centering
    \includegraphics[width=0.9\columnwidth]{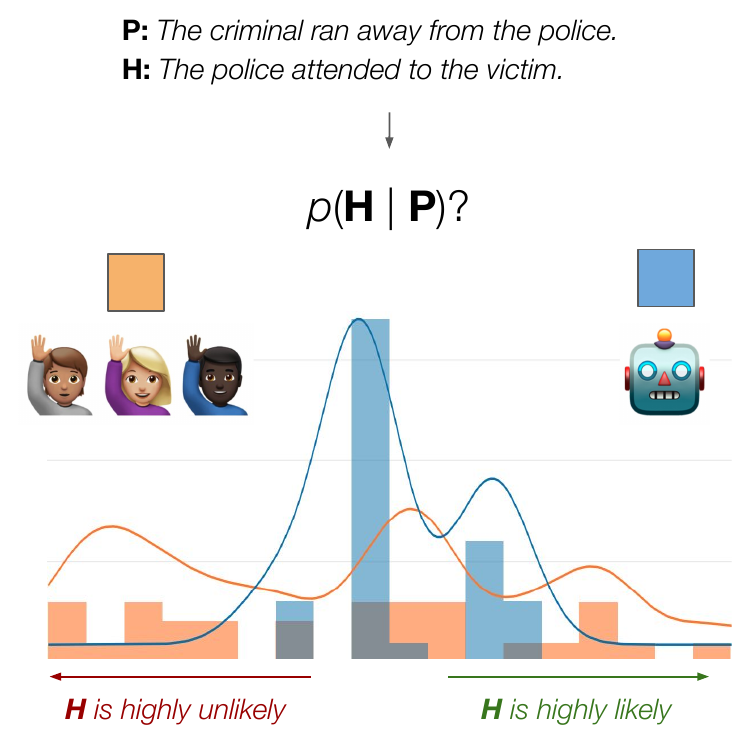}
    \caption{High-level overview of our paper. We use \textsc{ProbCOPA}, a novel dataset of probabilistic inferences, to collect judgments of inference likelihood from humans and models, and study how well their respective judgment distributions align with one another.}
    \label{fig:Figure-1}
\end{figure}

\section{Introduction}
\label{sec:introduction}
Much of the day-to-day reasoning that humans do involves working over partial information to arrive at probabilistic conclusions \citep{oaksford2007bayesian}. Consider:

\begin{exe}
    \ex \textit{There was an accident on the highway.} $\xrightarrow{}$ \textit{Traffic was worse than usual.}
    \label{ex:accident-mid-probability}
    \ex \textit{There was an accident on the highway.} $\xrightarrow{}$ \textit{Traffic was largely unaffected.}
    \label{ex:accident-low-probability}
\end{exe}

In the absence of any further context, (\ref{ex:accident-mid-probability}) and (\ref{ex:accident-low-probability}) involve conclusions that may or may not be true given the information presented.
Instead, the two conclusions are only likely or unlikely to varying degrees, given the first statement as well as background knowledge about highways and car accidents.
Although (\ref{ex:accident-mid-probability}) is likely, it is not guaranteed\textemdash perhaps everyone avoided the highway after hearing the news, leading to less traffic.
Conversely, although (\ref{ex:accident-low-probability}) is unlikely, it cannot be ruled out completely\textemdash maybe the vehicles involved were swiftly moved out of the way, leading to minimal impact on traffic.
We refer to such reasoning as \textit{probabilistic reasoning}, and individual inferences of the kind in (\ref{ex:accident-mid-probability}) and (\ref{ex:accident-low-probability}) as \textit{probabilistic inferences}.

What does such reasoning look like in humans and large language models (LLMs)?
In this paper, we provide initial insights into this question.
We compare probabilistic reasoning in humans and LLMs, in terms of their respective judgments towards a range of commonsense probabilistic inferences. 
Overall, we find that while models generally align with human judgments for probabilistic inferences deemed highly likely or highly unlikely, they consistently struggle to align with human judgments towards probabilistic inferences where annotators show more uncertainty (i.e., inferences that were deemed neither highly unlikely nor highly likely) and almost never show human-level judgment variation across sampled responses. We  make the following contributions:
\begin{itemize}
    \item We introduce \textsc{ProbCOPA}, a novel dataset of 210 handcrafted probabilistic inferences in English, with at least 25 human annotations per item.
    \item We highlight persistent differences between humans and LLMs in their judgments towards such probabilistic inferences.  
    \item We identify patterns in LLM reasoning chains that shed light on how they arrive at their final responses in these contexts.
\end{itemize}

\section{The \textsc{ProbCOPA} Dataset}
\label{sec:probcopa}

\subsection{Data Construction}
\label{subsec:data-source}
We aim to study inferences that are not strictly logically entailed, but rather those that lie on a range of likelihood given a premise.
Due to the limitations of existing NLI datasets in this regard (see \Cref{subsec:nli}), we construct our own corpus of probabilistic inferences.

We begin with the Choice of Plausible Alternatives (\textsc{COPA}) dataset \citep{roemmele2011choice}, which consists of 1,000 manually handcrafted items that probe commonsense reasoning.

\begin{exe}
    \ex \underline{Premise:} A drought occurred in the region. \textit{What happened as a result?}
    \label{ex:copa-sample}
    \begin{xlist}
        \ex Alternative 1: The crops perished.
        \label{ex:copa-sample-a1}
        \ex Alternative 2: The water became contaminated.
        \label{ex:copa-sample-a2}
    \end{xlist}
\end{exe}

As the example in (\ref{ex:copa-sample}) illustrates, each \textsc{COPA} item consists of a single premise together with two possible effects or causes.
For each pair of alternatives, one alternative is more plausible than the other; accordingly, the original task formulation involves choosing the \textit{more likely} alternative among the two ((\ref{ex:copa-sample-a1}) in this example).
Importantly, however, both alternatives are designed to be at least slightly plausible given the premise.

To construct our dataset, we therefore split each \textsc{COPA} item into two NLI-style items, such as (\ref{ex:copa-nli-a1}) and (\ref{ex:copa-nli-a2}):

\begin{exe}
    \ex {P:} A drought occurred in the region. \\
    {H:} The crops perished.
    \label{ex:copa-nli-a1}
    \ex {P:} A drought occurred in the region. \\
    {H:} The water became contaminated.
    \label{ex:copa-nli-a2}
\end{exe}

Crucially, although the alternatives are designed to yield clear preferences when evaluated against each other, when each is evaluated in isolation, they constitute probabilistic inferences with varying degrees of plausibility or likelihood.

Due to frequently-attested complexities in people's estimation of causal likelihood \citep{eddy1982inverse,villejoubert2002inverse,krynski2007role,stilgenbauer2017reasoning}, we exclude \textsc{COPA} items that ask participants to reason over possible causes, and only include those that elicit judgments on effect likelihood given a premise.
We take a random sample of 105 such items from the \textsc{COPA} test set and split each of these as described above, resulting in 210 probabilistic inferences framed as NLI-style datapoints.

\subsection{Human Annotation Procedure}
\label{subsec:human-annotation}
We conducted online crowdsourced experiments via Prolific\footnote{\url{https://www.prolific.com}} to obtain human annotations for our dataset.
We recruited 328 native English speakers based in the U.K., U.S. or Canada; these participants each annotated up to 30 \textsc{ProbCOPA} items under the procedure described below. 
All experimental protocols with humans were approved by our institution's Research Ethics Board, and participants were paid an average of US\$15.00/hr.

The annotation procedure involved crowdworkers being presented with one premise-hypothesis pair at a time, and rating the likelihood of the hypothesis as a result of the premise (using a sliding scale to return a numerical rating between 0 and 100).
Given attested variation in how humans express likelihood and uncertainty \citep{change2007intergovernmental, wintle2019verbal, ulmer2025anthropomimetic}, the sliding scale was shown to participants along with an aid suggesting how to partition values along it.

Participants began with five instructional examples for which they received feedback\textemdash this was meant to both explain the task format to them, as well as calibrate their responses within broad ranges of the numerical scale.
Following these examples, participants were presented with a sample of up to 30 test stimuli, with five attention checks interspersed in between.
The order of test stimuli was randomly shuffled for each participant, and all responses from participants who failed more than one attention check were discarded.

After discarding data from participants who failed the attention checks, we were left with between 25-30 likelihood score annotations (each from a unique participant) for each of our 210 items, with a median of 28 annotations per item. 

Appendix \ref{app:human-annotation} describes the human annotation set-up in further detail, and includes screenshots of the user interface used.

\subsection{Reproducibility of Human Responses}
\label{subsec:human-reproducibility}
We run two rounds of validation to ensure that our human responses are reproducible.
In the first, reported further in \Cref{subsec:model-results-methodology}, we have 30 items re-annotated by 30 new participants, and use this as a baseline for human-to-human response comparisons.
In the second round, we have the same 30 items re-annotated by another 30 new participants, but this time with a slightly different prompt wording.
We calculate the Spearman correlation between mean item ratings from our original annotations and each of these validation rounds; Spearman's $\rho=0.98$ ($p=4.52e-20$) and $0.97$ ($p=1.22e-19$) for the first and second validation rounds respectively.
Similarly, using two-sample Kolmogorov-Smirnov tests, we find no statistically significant differences in human response distributions under either of these conditions ($\alpha=0.05$).
Together, these validation results suggest our human annotations and strongly reproducible and trustworthy.

\section{Analysis of Human Responses}
\label{sec:human-results}

\subsection{Methodology}
\label{subsec:human-results-methodology}
\paragraph{On Normalizing Human Responses}
Studies that analyze human responses on a numerical scale often normalize human ratings (typically via by-participant $z$-scoring) to allow for comparisons across participants who may use the scale differently \citep[e.g.][]{sprouse2013comparison, mahowald2016snap, pavlick2019inherent}.
In this study, however, we deliberately instead work with the raw likelihood scores from participants, and analyze their distribution in relation to factors like human response time or their entropy.

We do so primarily for the following reason: since our scale explicitly corresponds to event likelihood, we often actually \textit{want} to preserve inter-annotator differences in scale use. 
For example, if an annotator only responds with values between 0 and 95, we argue that this means the annotator specifically chooses to never assign full likelihood (100) to an event, and that this behavioral pattern should be preserved in our analysis (rather than normalized away).
Moreover, given that participants received instructional feedback and guidance on how to use the scale prior to and during annotation, as well as being subject to attention checks (see \Cref{subsec:human-annotation}, Appendix \ref{app:human-annotation}), we trust that their responses are indications of their own likelihood judgments, rather than merely being artifacts of their scale use. 

\paragraph{Metric for Response Spread}
To quantify how spread out human responses are for each item, we use \textit{differential entropy}, an extension of Shannon entropy to continuous variables.
For a continuous random variable $X$ with probability density function $f(x)$, the differential entropy is defined as:

\begin{exe}
    \ex $h(X) = -\int f(x) \log f(x) dx$
    \label{formula:diff-entropy}
\end{exe}

Higher differential entropy values indicate greater dispersion in responses, while lower values indicate more concentration. 
We employ differential entropy over other metrics of spread (such as variance) due to its mathematical properties\textemdash intuitively, it captures the spread of information, rather than simply distance from the mean.
For instance, while a bimodal distribution with responses concentrated at either extreme of our scale would yield extremely high variance, its differential entropy would not be as high, as the information remains relatively tightly clustered (even if this is into two groups).
Crucially, however, unlike Shannon entropy, differential entropy can take negative values when a distribution is extremely concentrated, as we see for a handful of items in \Cref{fig:model-human-specific-joint}. 

\begin{figure*}
    \centering
    \includegraphics[width=\linewidth]{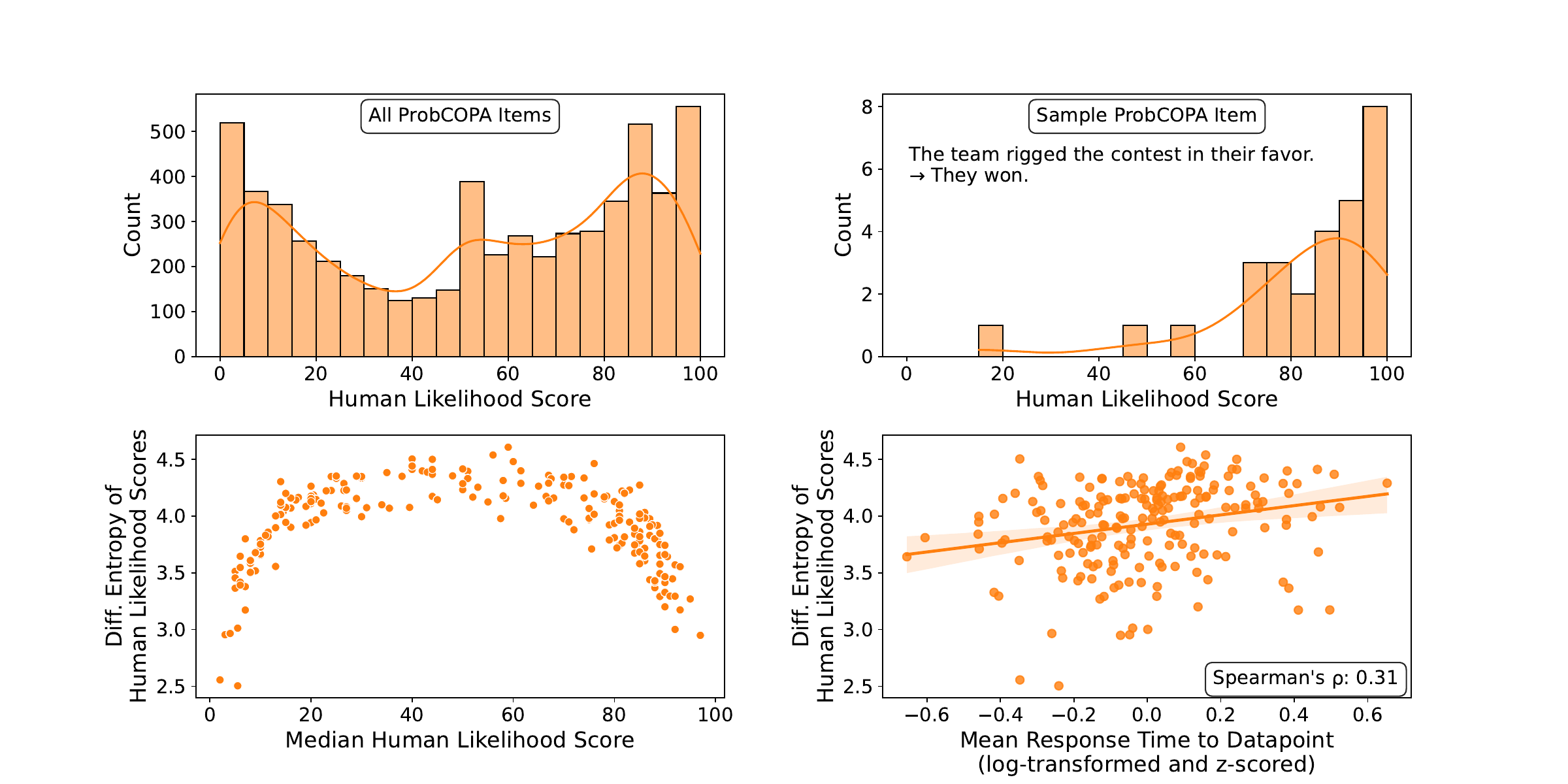}
    \caption{Distribution of human responses to \textsc{ProbCOPA}. \textbf{Top-right:} Likelihood scores across the entire dataset are tri-modal, with a significant proportion of responses between these modes; \textbf{Top-right:} likelihood scores for individual items typically follow a truncated normal distribution; \textbf{Bottom-left:} items with median responses towards extreme ends of the scale are subject to lower inter-annotator disagreement than for those in the middle ranges; \textbf{Bottom-right:} items with higher inter-annotator disagreement are (weakly) correlated with loinger response times from participants.}
    \label{fig:human-response-dist-joint}
\end{figure*}

\subsection{Results}
\label{subsec:human-results-results}

\paragraph{Likelihood scores from humans reveal graded, probabilistic judgments.}
\Cref{fig:human-response-dist-joint} (\textit{top-left}) shows the overall distribution of likelihood scores from human annotators, across the whole \textsc{ProbCOPA} dataset.
As it indicates, while the distribution of likelihood scores across the entire dataset shows three clear modes\textemdash corresponding to very low, very high, and balanced inference likelihood\textemdash a significant proportion of likelihood scores provided by annotators lie in between these modes, corresponding to more graded likelihood judgments.

In Appendix \ref{app:extended-results}, we show how this compares to similar human response data collected by \citet{pavlick2019inherent} on several major NLI datasets.
Like ours, the human responses collected by \citet{pavlick2019inherent} for these datasets are on a numerical scale, and correspond to the likelihood of hypotheses in NLI items.
Yet importantly, while they are also tri-modal, hardly any human responses lie between the three modes\textemdash indicating that the items in our dataset yield significantly more graded, probabilistic judgments than those in existing NLI datasets.

\paragraph{Human likelihood score distributions are almost always unimodal.}
While the overall distribution of likelihood scores across \textsc{ProbCOPA} is tri-modal, responses for \textit{individual} items are almost always unimodal.
\Cref{fig:human-response-dist-joint} (\textit{top-right}) shows the distribution of human likelihood scores for one such item in our dataset.
As it indicates, these responses approximate a Beta distribution with a single mode; we find that human responses for most items across the dataset do so as well.
To confirm that our data is indeed unimodal, we use \citeauthor{silverman1981using}'s \citeyearpar{silverman1981using} statistical test of multimodality.
The null hypothesis is that the sample distribution is unimodal; it is not rejected for any item in our dataset (at $\alpha=0.05$).

%%%%%%%%%%%%%%%%%%%%%%%%%%%%%%%%%%%%%%%%%%%%%%%%%%
%%%%%%%%%%<MODEL RESPONSE DISTRIBUTIONS> %%%%%%%%%
%%%%%%%%%%%%%%%%%%%%%%%%%%%%%%%%%%%%%%%%%%%%%%%%%%
\begin{figure*}
    \centering
    \includegraphics[width=\linewidth]{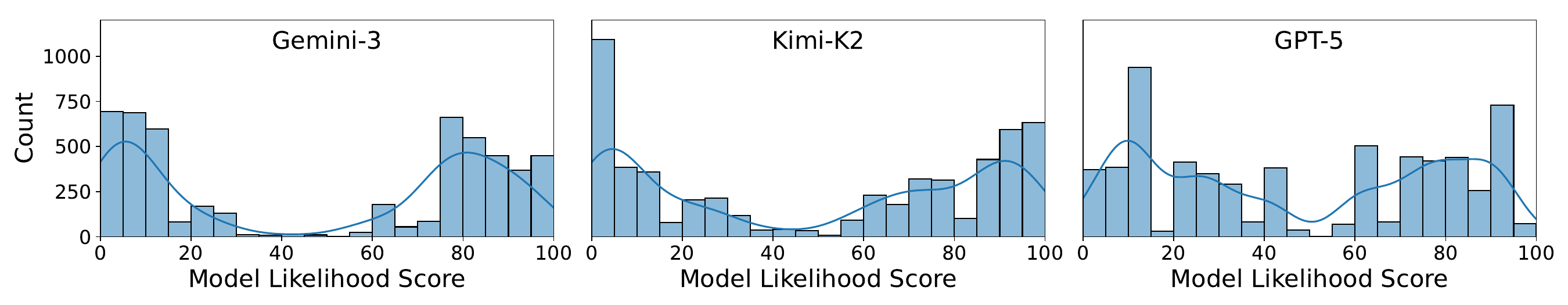}
    \caption{Distribution of likelihood scores across all \textsc{ProbCOPA} items, from three models. In contrast to humans (see \Cref{fig:human-response-dist-joint}), models rarely return responses indicating medium likelihood, though this tendency is less extreme with GPT-5. See \Cref{fig:model-response-distributions-all} for the full set of distributions by model.}
    \label{fig:model-response-distributions}
\end{figure*}
%%%%%%%%%%%%%%%%%%%%%%%%%%%%%%%%%%%%%%%%%%%%%%%%%%
%%%%%%%%% </MODEL RESPONSE DISTRIBUTIONS> %%%%%%%%%%
%%%%%%%%%%%%%%%%%%%%%%%%%%%%%%%%%%%%%%%%%%%%%%%%%%

The items in \textsc{ProbCOPA} therefore further stand out from those in existing NLI datasets, because while they are subject to judgment variation around an underlying mode, the fact that this variation is unimodal\textemdash rather than multimodal\textemdash suggests that the items are not subject to qualitative differences in interpretation, as has been previously reported for NLI datasets \citep{pavlick2019inherent, jiang2023ecologically}.

\paragraph{Annotators do not collectively agree on a hypothesis having medium likelihood.}
\Cref{fig:human-response-dist-joint} (\textit{bottom-left}) shows, for each item in our dataset, the differential entropy of human likelihood scores for the item ($y$-axis) plotted against its median likelihood score ($x$-axis).
The differential entropy of human likelihood scores follows a horseshoe-like shape\textemdash items receiving high or low median likelihood scores are associated with comparatively lower differential entropy (i.e., higher inter-annotator agreement), while items with median likelihood scores closer to the middle of the scale are associated with higher differential entropy (i.e., lower inter-annotator agreement).
Notably, we find no items for which annotators closely agree on a hypothesis having medium likelihood.

\paragraph{Higher entropy items are (weakly) correlated with longer human response times.}
\Cref{fig:human-response-dist-joint} (\textit{bottom-right}) shows the differential entropy of likelihood scores for each datapoint ($y$-axis), plotted against the mean time taken by participants to respond to that datapoint (log-transformed and $z$-scored; $x$-axis).
We find a positive correlation between response time and likelihood score entropy (Spearman's $\rho=0.31$, $p=6.45e-06$)\textemdash meaning that, on average, items yielding lower inter-annotator agreement were also items that participants took longer to respond to in our experiment.
We take this as evidence that the higher inter-annotator variation for some items is not a result of noise, but instead relates to item difficulty.

\section{Comparison with Responses from Reasoning LLMs}
\label{sec:model-responses}
Having analyzed how humans judge the probabilistic inferences in \textsc{ProbCOPA}, we now turn to LLMs.
In particular, we test \textit{reasoning LLMs}: LLMs that are trained to produce intermediate tokens (commonly referred to as a \textit{reasoning chain}) before outputting a final response \citep{xu2025toward, li2025system, marjanovic2025deepseek}.
We specifically focus on these models as (i) they represent the state-of-the-art on reasoning tasks, but (ii) are generally not evaluated on open-ended, non-deterministic reasoning contexts (see \Cref{sec:related-work}).

\subsection{Methodology}
\label{subsec:model-results-methodology}

\paragraph{Model Response Format}
We seek to obtain likelihood scores from reasoning models to compare with those we obtained from humans.
While previous work has used model log-probabilities or sigmoid/softmax distributions in similar contexts \citep{pavlick2019inherent, chen2020uncertain, kauf2024log}, these are not accessible for most state-of-the-art reasoning LLMs.
Moreover, since reasoning chains determine these models' final outputs, simply observing model probabilities conditioned on an input may fail to reflect actual output distributions when intermediate reasoning chains are generated.
Conversely, while uncertainty quantification methods for black-box models may offer inspiration \citep[e.g.][]{kuhn2023uncertainty, lin2024blackbox, ulmer2024calibrating}, these either (i) require gold labels against which accuracy can be measured, or (ii) are suited to open-ended generation settings.
But probabilistic inferences by definition do not involve hard labels against which accuracy is a meaningful metric, and our setting involves likelihood estimates rather than open-ended generations.

For these reasons, following \citet{mei2025reasoning}, we obtain likelihood scores from reasoning LLMs via verbalized numerical estimates.
For each item in our dataset, we ask the model to reason about the premise and hypothesis, and then return a value between 0 and 100 indicating the likelihood of the hypothesis given the premise.
When doing so, we also provide the model with the same guide provided to humans describing how to partition the numerical scale (see \Cref{subsec:human-annotation}).
We repeat this 30 times for each item under the models' default temperature settings, to sample 30 likelihood scores per item for each model.
The full prompt we provide to models is presented in Appendix \ref{appendix:prompt}.

\paragraph{Metric for Distributional Comparison}
We again use differential entropy to quantify the spread of model likelihood scores (see \Cref{subsec:human-results-methodology}). 
To make distributional comparisons between human and model scores, however, we need a metric of distributional similarity.
We use \textit{Wasserstein distance} (also known as \textit{Earth Mover's Distance}). 
Formally, for two distributions $P$ and $Q$, this is defined as:

\begin{exe}
    \ex $W_1(P, Q) = \inf_{\gamma \in \Gamma(P,Q)} \mathbb{E}_{(x,y) \sim \gamma}[|x - y|]$
    \label{ex:wasserstein-formula}
\end{exe}
where $\Gamma(P,Q)$ denotes the set of all joint distributions with marginals $P$ and $Q$. 
Intuitively, this captures the `cost' of transforming one probability distribution into the other; higher values indicate lower distributional similarity, and lower values indicate higher similarity.\footnote{We use this measure of distributional similarity over $KL$-divergence (another popular metric of distributional divergence), because unlike the latter, it does require the distributions to have matching support\textemdash model and human responses need not cover the same ranges of the likelihood scale.}

\begin{figure*}
    \centering
    \includegraphics[width=0.95\linewidth]{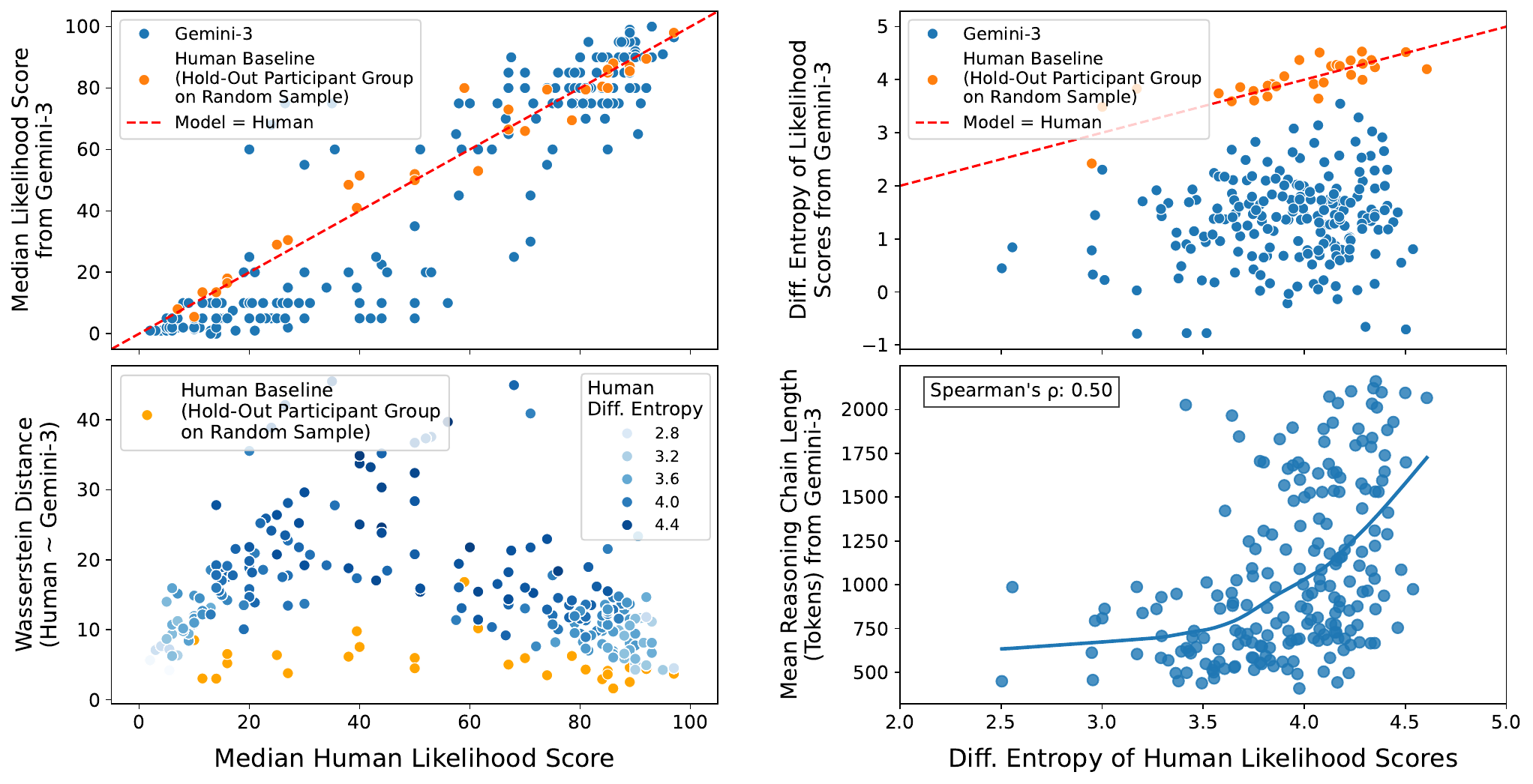}
    \caption{Item-wise comparisons between Gemini-3 and humans. \textbf{Top-left:} median likelihood scores from Gemini-3 align with those from humans at extreme ends of the scale, but not in the middle ranges; \textbf{Bottom-left:} likelihood score distributions from Gemini-3 and humans reflect the same pattern, with highest divergences for middle-range items (which also saw less inter-annotator agreement); \textbf{Top-right:} Gemini-3 shows less response diversity that humans for all items; \textbf{Bottom-right:} Gemini-3 on average reasons longer for items that humans disagree more on.}
    \label{fig:model-human-specific-joint}
\end{figure*}

\paragraph{Models Tested}
We test a range of contemporary reasoning LLMs from different model providers: Gemini-3 \citep{google2025gemini3}, GPT-5 \citep{openai2025gpt5}, Claude Sonnet-4.5 \citep{anthropic2025sonnet45}, Qwen3 \citep{qwen32025}, Kimi-K2 \citep{kimiTeam2025kimik2}, GLM-4.6 \citep{zai2025glm45, zai2025glm46}, DeepSeek-R1 \citep{deepseekr1}, and Grok-4.1 Fast \citep{xai2025grok41}.
For details on model versions and how we ran inference, see Appendix \ref{appendix:model-details}.

As a follow-up, we also run preliminary experiments with Claude Opus-4.6 \citep{anthropic2026opus46}, but find that this model returned almost completely deterministic responses for each item, without providing any reasoning chains.
We discuss these results in Appendix \ref{app:claude-opus-4-6}, but exclude them from our main analysis, as it remains unclear how informative they actually are.

\paragraph{Human Baseline}
When evaluating how closely model responses align with human likelihood scores, we also want a baseline of how well other humans can approximate these same scores.
To establish this baseline, we therefore have a random sample of 30 \textsc{ProbCOPA} items re-annotated by a fresh set of participants, under the same annotation procedure reported in \Cref{subsec:human-annotation}. 
When comparing a reasoning LLM's likelihood scores with those of \textsc{ProbCOPA} annotators, we then use this hold-out participant group's annotations to compute a baseline for human-to-human response similarity.

\subsection{Results}
\label{subsec:model-results-results}

\paragraph{Models rarely indicate medium likelihood.}
\Cref{fig:model-response-distributions} shows the overall distribution of likelihood scores (across all \textsc{ProbCOPA} items) for Gemini-3, Kimi-K2, and GPT-5. 
As it demonstrates, models exhibit a tendency not to return likelihood scores in the middle of the scale (i.e., those indicating medium likelihood).  
Though this tendency is least extreme for GPT-5 (see \Cref{fig:model-response-distributions-all} in Appendix \ref{app:extended-results} for the full set of model likelihood score distributions), it too rarely returns values in the very middle of the scale.
Models thus appear committed to strong judgments of inference likelihood, supporting prior findings that they are often overconfident (see \Cref{subsec:uq-for-llms}).

\paragraph{Model responses align with human responses more for low- and high-likelihood items than those in between.}
\Cref{fig:model-human-specific-joint} (\textit{top-left}) shows the median human likelihood score ($x$-axis) against the median score from Gemini-3 ($y$-axis) for each \textsc{ProbCOPA} item.
As it suggests, while median responses from the model are similar to those from humans at the two extreme ends of the scale, this relationship breaks down closer to the middle of scale (since, as mentioned, models avoid responses in this range).
As our baseline indicates, however, other humans are capable of reproducing similar median judgments for items across the scale. 

We find this trend also holds when comparing entire distributions.
\Cref{fig:model-human-specific-joint} (\textit{bottom-left}) shows, for the same model, item-wise Wasserstein distances ($y$-axis) between human and model responses, as a function of the median likelihood score ($x$-axis) and differential entropy (color) from humans.
As the plot suggests, distributional similarity between model and human likelihood scores is highest for items that humans collectively deem highly likely or unlikely, and lowest for items without such a consensus.
Once again, however, we find no such pattern in our human baseline, which shows roughly the same degree of distributional similarity between our original and subsequent baseline annotations across all items. 
These trends hold for all models tested; full results are shown in \Cref{fig:model-human-wasserstein-distance-all,fig:model-human-median-responses-all} (Appendix \ref{app:extended-results}).

\paragraph{Model responses almost never show as much variation as human responses.}
\Cref{fig:model-human-specific-joint} (\textit{top-right}) compares, at an item-wise level, the differential entropy of likelihood scores from Gemini-3 ($y$-axis) and our original human annotators ($x$-axis).
As it shows, for every single item in our dataset, likelihood scores from humans show higher differential entropy (in other words, more variation) than those from Gemini-3.
Comparing with our human baseline, however, reveals roughly similar response variation between participants.
Results for the full set of models tested are presented in \Cref{fig:model-human-entropy-all} in Appendix \ref{app:extended-results}; models almost never show more variation in their responses than humans.

We run follow-up experiments with the same random sample of 30 items we use for our human baseline, and find that adjusting temperature does not yield human-level response entropy from models; when increasing temperature, models will often devolve to generating endless sequences of random tokens before ever achieving human-like response variability.
Similarly, we find that persona prompting \citep{zheng-etal-2024-persona, luz-de-araujo-etal-2025-persona} similarly has limited effects, and fails to deliver human-level response variation. 
These results are shown in \Cref{fig:unified-temperature-ablation,fig:unified-persona-prompting-ablation}, and support the general finding that state-of-the-art models often struggle to represent human variation \citep{santurkar2023whose, zhang2025cultivating}.

%%%%%%%%%%%%%%%%%%%%%%%%%%%%%%%%%%%%%%%%%%%%%%%%%%
%%%%%%%%%% <ENSEMBLE WASSERSTEIN BOXPLOT> %%%%%%%%
%%%%%%%%%%%%%%%%%%%%%%%%%%%%%%%%%%%%%%%%%%%%%%%%%%
\begin{figure*}
    \centering
    \includegraphics[width=\linewidth]{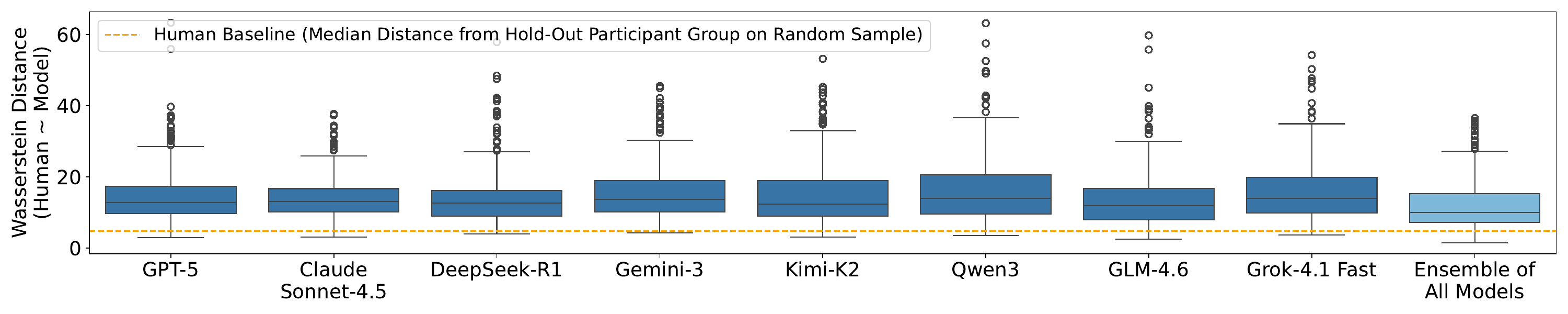}
    \caption{Distribution of item-wise Wasserstein distances between human and model likelihood score distributions. Ensembling the outputs of all models yields better distributional alignment with human judgments, but still falls short of the human-human baseline.}
    \label{fig:ensemble-wasserstein-distance-boxplot}
\end{figure*}
%%%%%%%%%%%%%%%%%%%%%%%%%%%%%%%%%%%%%%%%%%%%%%%%%%
%%%%%%%%% </ENSEMBLE WASSERSTEIN BOXPLOT> %%%%%%%%
%%%%%%%%%%%%%%%%%%%%%%%%%%%%%%%%%%%%%%%%%%%%%%%%%%

\paragraph{Increased reasoning effort does not significantly change models' likelihood scores.}
In the same follow-up study, we also assess whether increasing reasoning LLMs' \textit{reasoning effort} parameters lead to significantly different outputs.
For each item in human baseline random sample, we compare the median likelihood score produced under `low' and `high' reasoning effort parameters\footnote{For Gemini-3 and Claude Sonnet-4.5, we use thinking budgets of 512 and 4096 respectively to simulate the contrast between `low' and `high' reasoning effort.}, and use bootstrapped $95\%$ confidence intervals for these medians to check for statistical significance.
Across all 30 items in the sample, and across all models, we never find any case of higher reasoning effort leading to a statistically significant difference in median likelihood score.
Notably, these findings contrast those from \citet{mei2025reasoning}, who find that increased reasoning yields more overconfidence.
See \Cref{fig:unified-reasoning-effort-ablation} for further details.

\paragraph{Ensembling model responses leads to better (but not human-like) alignment with human response distributions.}
\Cref{fig:ensemble-wasserstein-distance-boxplot} shows, for each model tested, the distribution of item-wise Wasserstein distances between model and human likelihood score distributions.
As it demonstrates, model-human distributional differences are far higher than human-human distributional differences; ensembling model responses reduces this gap, but still falls short of the human baseline.

\section{Analyzing LLM Reasoning Chains}
\label{sec:reasoning-chain-analysis}
Finally, we study reasoning LLMs' reasoning chains, to identify common patterns in how they reason over probabilistic inferences.

\paragraph{Models reason longer for items that humans disagree more about...}
\Cref{fig:model-human-specific-joint} (\textit{bottom-right}) shows, for each \textsc{ProbCOPA} item, the mean reasoning chain length from Gemini-3 ($y$-axis; measured in tokens) against the differential entropy of human likelihood scores for that same item ($x$-axis).
We see a clear correlation (Spearman's $\rho=0.50$, $p=2.10e-14$) between reasoning chain length and human differential entropy: on average, items that humans showed more uncertainty on yielded longer LLM reasoning chains. 
\Cref{tab:reasoning-chain-correlations,fig:reasoning-chain-length-human-diff-entropy-all} in Appendix \ref{app:extended-results} show these results for all models tested: most show at least modest correlations ($\rho \geq 0.30$), even if these are weaker than for Gemini-3.

\paragraph{...but correlations with human response time are much weaker.}
Despite this, we find that correlations between reasoning chain length and human response time (log-transformed and by-participant $z$-scored) are far lower.
\Cref{tab:reasoning-chain-correlations,fig:reasoning-chain-length-human-response-time-all} show these correlations for all models tested.
While the highest correlation achieved between reasoning chain length and human differential entropy is $0.50$ (from Gemini-3), the highest such correlation between reasoning chain length and human response time is only $0.25$ (from Qwen-3).
This indicates that while reasoning chain lengths may carry some relationship with how human judgments are distributed, relations to human cognitive load are far less clear. 

\paragraph{Models explicitly reason over alternatives to arrive at likelihood judgments.}
For more qualitative insights into how models reason over \textsc{ProbCOPA} items, we manually inspect a random sample of 100 reasoning chains across all model responses.
Doing so, we find a common pattern: 90 out of the 100 reasoning chains sampled include explicit considerations of alternative scenarios that are used to frame the model's final response. 
\Cref{tab:reasoning-chain-alternatives} shows some examples of this pattern, across different models.
Though subject to questions of faithfulness \citep[see][]{lanham2023measuring, xiong2025measuring, chen2025reasoning}, the consistent use of alternative scenarios points to a common reasoning pattern across models, and invites further questions into how well this aligns with humans.

\sethlcolor{yellow}
\begin{table*}[htbp]
\centering
{\footnotesize
\begin{tabular}{>{\raggedright\arraybackslash}p{0.1\textwidth}
                >{\raggedright\arraybackslash}p{0.1\textwidth}
                >{\raggedright\arraybackslash}p{0.1\textwidth}
                >{\raggedright\arraybackslash}p{0.55\textwidth}}
\toprule
\footnotesize Model & \footnotesize Premise & \footnotesize Hypothesis & \footnotesize Reasoning Chain Extract \\
\midrule
\fontsize{7.5}{9}\selectfont Kimi-K2-Thinking & \fontsize{7.5}{9}\selectfont The tire on my bike was flat. & \fontsize{7.5}{9}\selectfont I pumped air into the tire. & \fontsize{7.5}{9}\selectfont ...When a bike tire is flat, the most common and logical action is to pump air into it... However, \hl{there are other possibilities}: ... \hl{They might not fix it at all}... \hl{They might replace the tube/tire entirely}... \hl{they might take it to a shop}...\\
\midrule
\fontsize{7.5}{9}\selectfont DeepSeek-R1 & \fontsize{7.5}{9}\selectfont The girl looked pale. & \fontsize{7.5}{9}\selectfont Her father read her a story. & \fontsize{7.5}{9}\selectfont ...This is an action a parent might take to comfort a child... \hl{However, it's not the only response. The father might giver her medicine, ask what's wrong, take her to a doctor, or just let her rest}... \\
\midrule
\fontsize{7.5}{9}\selectfont GLM-4.6 & \fontsize{7.5}{9}\selectfont The man anticipated the team's victory. & \fontsize{7.5}{9}\selectfont He made a bet with his friends. & \fontsize{7.5}{9}\selectfont ...when people anticipate a victory for a team they support, they might be more inclined to bet on that outcome... \hl{However}... \hl{They might not be gamblers}... \hl{They might not have money to bet}... \hl{They might be satisfied with just watching the game}... \\
\bottomrule
\end{tabular}
}
\caption{Sample excerpts of reasoning chains from different models, demonstrating the explicit considerations of alternative outcomes of the premise (highlighted in \hl{yellow}).}
\label{tab:reasoning-chain-alternatives}
\end{table*}

\section{Discussion}
\label{sec:discussion}
Our results offer initial insights into how models reason in open-ended, non-deterministic settings, and point to the potential of further research in this area.
For instance, our findings indicate that the tendency for models to be overconfident in their outputs \citep{mielke2022reducing, mei2025reasoning, tian2023calibration} is reflected even in open-ended inferences that are inherently uncertain; the models we test rarely indicate medium likelihood, and instead consistently favor more extreme likelihood scores.
Similarly, our experiments reveal persistent differences between humans and models, with models failing to closely align with human judgment distributions, and producing far less variation in their responses than humans, even with different temperature settings (see \Cref{subsec:model-results-results}).
Such issues of human-model similarity are of increasing importance as LLMs are used in human-focused settings \citep[see e.g.][]{maity2025large, wilcox2025bigger, anthisposition2025}, and our work reiterates the need to assess models vis-\`{a}-vis these comparisons.

Conversely, our findings also carry relevance for studies of human reasoning.
Most notably, the graded, probabilistic judgments we see from our study participants (see \Cref{subsec:human-results-results}) serve as empirical evidence of the probabilistic aspects of human reasoning and inference \citep{oaksford2007bayesian}.  
Likewise, the observation that our human judgment distributions are unimodal stands out from recent work finding significant (often bimodal) human judgment variation towards NLI data \citep{pavlick2019inherent, nie2020chaosnli, jiang2023ecologically}, and suggests that sharp divergences in human inference judgments may not arise if the correct data is used \citep[see also][]{jiang2022investigating}.

\section{Related Work}
\label{sec:related-work}

\paragraph{Reasoning in Humans}
\label{subsec:reasoning-in-humans}
Foundational work in modern mathematics and linguistics characterized inference patterns in mathematics and natural language vis-\`{a}-vis formal logic \citep{frege1879begriffsschrift, tarski1936wahrheitsbegriff, montague1970universal}.
Empirical research in psychology, however, has since suggested these are not the kinds of reasoning patterns humans actually demonstrate, with several studies pointing to recurrent logical `fallacies' from humans (e.g.\ \citealt{wason1968reasoning, evans1983conflict, evans1999reasoning, klauer2000belief}, see \citealt{evans2002logicreview} for a review). 
Most notably, \citet{wason1968reasoning} demonstrated that humans frequently make faulty inferences from simple conditional statements, while \citet{evans1983conflict} showed humans frequently accept logically invalid arguments if their conclusions are believable.
\citet{oaksford2007bayesian} thus argue that human reasoning should instead be understood in terms of probabilistic beliefs\textemdash a motivation we operationalize in this study.

\paragraph{Natural Language Inference}
\label{subsec:nli}
In NLP, textual inferences are most often formalized via the \textit{natural language inference} (NLI) task.
Given a premise $P$ and hypothesis $H$, the task traditionally involves classifying the sentence pair as having an \textit{entailment}, \textit{contradiction} or \textit{neutral} relation \citep{dagan2005pascal}.\footnote{Note that entailment and contradiction in NLI typically refer to the notion that the hypothesis is \textit{most likely} true/false given the premise, as opposed to logically entailed/contradicted by it. See \citet{zaenen2005local,manning2006pascal,crouch2006circumscribing} for more discussion.}

NLI has been used to study both specific types of inferences in NLP systems \citep[e.g.][]{chen2020uncertain, bhagavatula2020abductive, tian2021diagnosing, liu2023we, zhang2017ordinal, jeretic2020imppres} as well as general natural language understanding \citep[see][]{poliak-2020-survey, madaan2025lost}.
A growing body of research, however, reveals significant human judgment variation in NLI tasks \citep{demarneffe2012did, pavlick2019inherent, nie2020chaosnli, jiang2022investigating, jiang2023ecologically, weber2024varierr}.
Crucially, this line of work indicates that items in popular NLI datasets, such as \textsc{SNLI} \citep{bowman2015large} and \textsc{MNLI} \citep{williams2018broad}, are subject to judgment variation that goes beyond noise or crowdworker errors \citep{pavlick2019inherent,weber2024varierr}.

Closest to our work, \citet{chen2020uncertain} re-annotate the \textsc{SNLI} dataset using a probabilistic scale, to study NLI vis-\`{a}-vis probabilistic inferences.
While their work is thus similar to ours in motivation, the data they use prevents the kind of analysis we conduct.
Most notably, as \citet{nighojkar2023no} note, the authors use the mean of only 2-3 crowdworker annotations as the gold label for each item.
But given that \textsc{SNLI} items have been shown to yield bimodal judgment distributions \citep{pavlick2019inherent}, these averages may be misleading; if one annotator judges an inference to be highly likely, and the other judges it to be highly unlikely, the resulting mean would indicate medium likelihood, even when no annotator believes this.
These limitations motivate us to construct and annotate our own dataset, which we detail in \Cref{sec:probcopa}. 

\paragraph{Reasoning LLMs}
\label{subsec:reasoning-llms}
Reasoning LLMs, like OpenAI's o3 \citep{openaio32025} or DeepSeek AI's DeepSeek-R1 \citep{deepseekr1}, are trained to produce intermediate tokens (known as a \textit{reasoning chain} or \textit{thinking trace}) before outputting a final response \citep{xu2025toward, li2025system, marjanovic2025deepseek}.\footnote{The use of terms like `reasoning' and `thinking' to describe these models has led to criticism from some that this anthropomorphizes LLMs \cite{kambhampati2025stop}. Here, we follow the common convention of the field to refer to such models as ``reasoning LLMs''. We do not, however, aim to imply that reasoning chains are akin to human thoughts.}
These LLMs appear to have induced strong reasoning capabilities, showing strong gains on several code and reasoning benchmarks \citep{openaio12024, deepseekr1, team2025kimi, yang2025qwen3, liu2025reinforcement}.

Crucially, however, this work is largely centered around mathematical and logical reasoning.
The reinforcement learning pipelines used by reasoning LLMs typically involve training on math or coding tasks that are automatically verifiable \citep{lambert2024tulu, liu2025reinforcement}.
Similarly, in evaluation, math and coding benchmarks like \textsc{AIME} \citep{aime2024} and \textsc{SWE-Bench} \cite{jimenezswe} are frequently used to make assessments of these LLMs' reasoning capabilities \citep[see e.g.][]{deepseekr1, openaio12024, openaio32025, team2025kimi, yang2025qwen3}.
As a result, little work has explored how reasoning LLMs behave in reasoning contexts that are more open-ended and non-deterministic.

Some work \textit{has} looked at how LLMs reason with probabilities \citep[][]{renda2025openestimate,pournemat2025reasoning,paruchuri2024odds,xia2024let,nafar2025extracting}, finding mixed results in terms of these abilities.
But importantly, such work frames `probabilistic reasoning' as correctly applying probability theory or inducing explicit statistical distributions (e.g.\ \textit{What is the percentile of 294mm precipitation?}).
Ours, on the other hand, focuses on reasoning over everyday, uncertain events, without requiring the reasoning process to involve explicit math or probability theory (see e.g. \Cref{tab:reasoning-chain-alternatives}).

\paragraph{Uncertainty Quantification for LLMs}
\label{subsec:uq-for-llms}
Finally, our work bears some relevance to uncertainty quantification (UQ) for LLMs. 
UQ in the context of LLMs asks how certain or confident models are of their outputs, typically in contrast to some measure of how certain or confident they \textit{should} be \citep[][]{ulmer2024uncertainty, liu2025uncertainty, shorinwa2025survey}.

Since we are only interested in how likely or unlikely LLMs deem some probabilistic inference to be\textemdash rather than how much uncertainty a model shows around any such probability estimate\textemdash our work is slightly outside the scope of traditional UQ methods \citep[see][]{lin2024blackbox}.
Nevertheless, to the extent that probabilistic inferences are by definition uncertain and non-deterministic, some work in UQ is highly relevant to our study.

For instance, several studies have examined how LLMs explicitly verbalize uncertainty, both through numerical estimates or linguistic markers \citep[][see \citealt{ulmer2025anthropomimetic} for an overview]{lin2022teaching, yona2024can, tian2023calibration, belem2024perceptions}.
Much of this line of work finds that LLMs are \textit{overconfident}, often being more confident of their outputs than is warranted \citep{mielke2022reducing, tian2023calibration, krause2023confidently, mei2025reasoning}.
Closest to our study, \citet{mei2025reasoning} find that reasoning LLMs are typically overconfident, and that deeper reasoning from models leads to greater overconfidence. 

\section{Conclusion}
\label{sec:conclusion}
In this paper, we assessed probabilistic reasoning in both humans and LLMs, using \textsc{ProbCOPA}, a novel dataset of 210 probabilistic inferences in English, each with at least 25 human annotations.
We find significant differences between how humans and reasoning LLMs judge probabilistic inferences, with models failing to match human judgment distributions or produce human-level output variation.
Furthermore, we analyze model reasoning chains, and identify common reasoning patterns, but mixed correlations with human behavior.
We hope our work inspires further research on reasoning beyond logical or deductive reasoning, and in more open-ended, human-like and non-deterministic contexts.

\section*{Limitations}
\label{sec:limitations}
Besides being limited to English, our study is subject to other limitations we highlight below.

\paragraph{Verbalized Likelihood Scores} While we argue that reasoning LLMs are best-suited to verbalized likelihood scores for the purposes of our study (see \Cref{subsec:model-results-methodology}), questions remain around how faithful these generally are \citep{tian2023calibration, kumar2024confidence}.
We thus hope that future work identifies other methods for likelihood elicitation that are suited to the specific nature of reasoning models.

\paragraph{\textsc{COPA}-Derived Items} Our dataset is novel, but its items are derived from \textsc{COPA} \citep{roemmele2011choice}, an older dataset that likely features in the training data of most models. 
Since our re-framing of the task around these items yields new judgments compared to the original \textsc{COPA} gold labels (see \Cref{subsec:data-source}), we do not believe that we are testing on a task the model has already been trained on; nevertheless, it is possible that the presence of the some of these sentences in the training data affects model behavior towards them.

\section*{Acknowledgments}
The authors would like to thank Verna Dankers, Marius Mosbach, Dennis Ulmer, Ivan Titov and Desmond Elliot for providing crucial feedback on this work.
This work was also made possible with the support of the IVADO \textit{R3 NLP Régroupement}, the Canada CIFAR AI Chair and the NSERC Discovery Grant.  
Gaurav Kamath is supported by a Doctoral Training Award from the \textit{Fonds du R\'{e}cherche du Qu\'{e}bec -- Soci\'{e}t\'{e} et Culture}.
Marie-Catherine de Marneffe is a Research Associate of the \textit{Fonds de la Recherche Scientifique -- FNRS}.
Sebastian Schuster has been supported by the Vienna Science and Technology Fund (WWTF) [10.47379/VRG23007] \textit{Understanding Language in Context.}
Every word in this paper was written by a human. 

\bibliography{main}

\appendix

\section{\textsc{ProbCOPA} Human Annotation Procedure}
\label{app:human-annotation}
\begin{figure*}
    \centering
    \includegraphics[width=\linewidth]{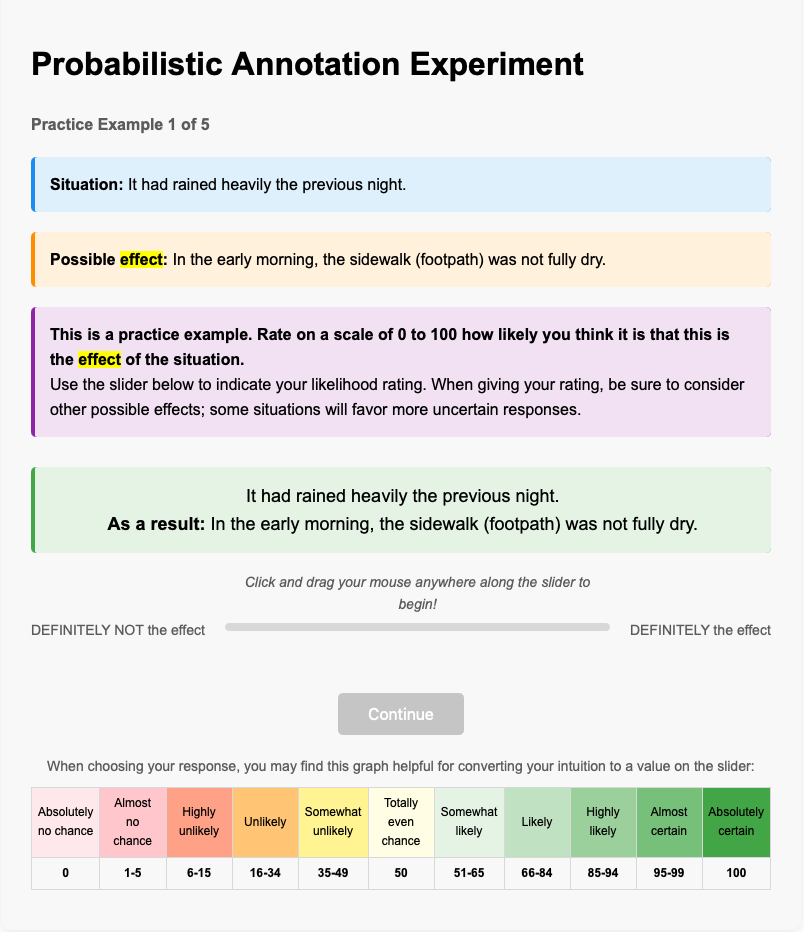}
    \caption{Screenshot of the first instructional example presented to participants in our crowdsourced experiment. Participants were shown the general task format, and asked to present a response using the slider. The guide presented beneath the example was intended to align participants on how to use the scale. In this instructional stage, participants were given automatic feedback based on their responses.}
    \label{fig:human-annotation-instructions-example}
\end{figure*}

\begin{figure*}
    \centering
    \includegraphics[width=\linewidth]{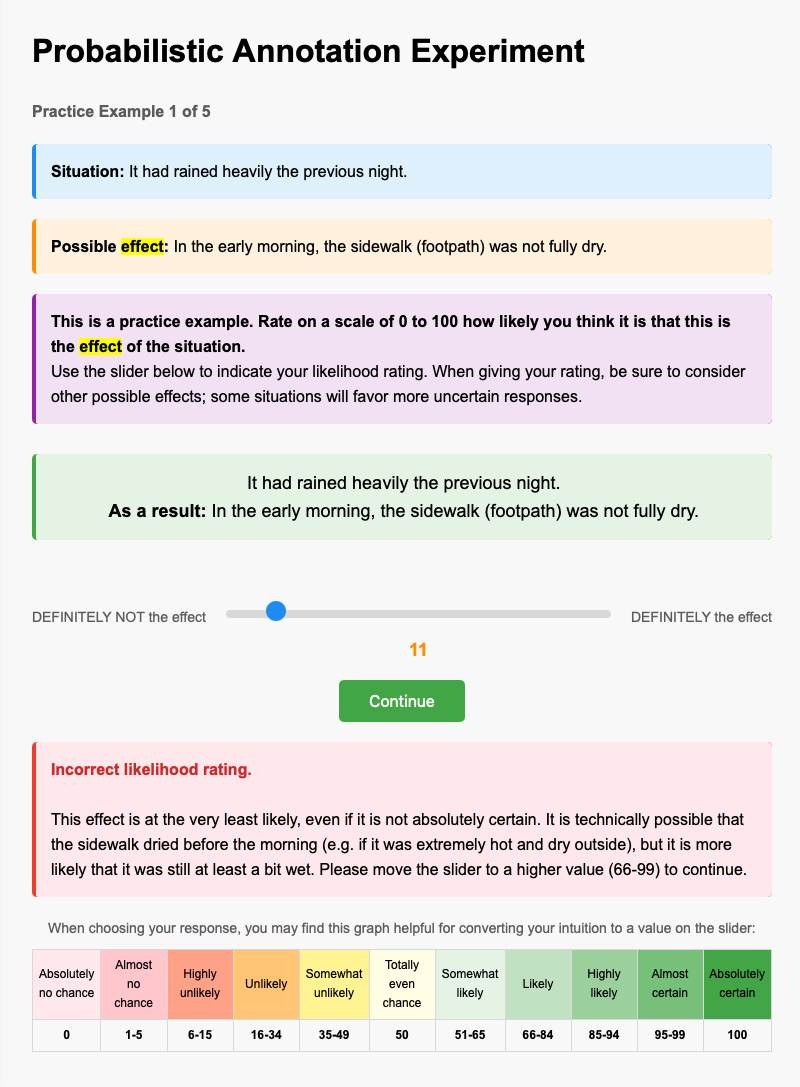}
    \caption{Screenshot showing the automatic feedback participants would receive if they provided a likelihood score outside of the `likely' to `almost certain' range for the first instructional example (see \Cref{fig:human-annotation-instructions-example}). In this stage, we aimed to use simple examples for which most people would agree on broad likelihood ranges.}
    \label{fig:human-annotation-instructions-example-wrong}
\end{figure*}

Human annotators recruited via Prolific participated in our crowdsourced experiment that we ran using a custom website built on HTML and JavaScript.
After reviewing and accepting a consent form, participants were presented with instructional examples that demonstrated the task format.
\Cref{fig:human-annotation-instructions-example} shows the first such instructional example: participants are presented with the general task format, and requested to provide a response using the slider. 
Note that we framed hypotheses as \textit{possible effects} of premises due to the original setting of \textsc{COPA} items, as well as because doing so offers an intuitive interpretation of inference likelihood.
Upon submitting a response, they would receive automatic feedback based on the range in which they responded.
In this phase, we aimed to use simple examples for which most people would share a broad consensus on likelihood ranges. 
\Cref{fig:human-annotation-instructions-example-wrong} shows an example of such feedback, if participants provided the `wrong' response (participants would also receive positive feedback if they responded within the intended ranges of the scale).
Participants were presented with 5 such instructional examples that familiarized them with low, middle, and high ranges of the scale.

\begin{figure*}
    \centering
    \includegraphics[width=\linewidth]{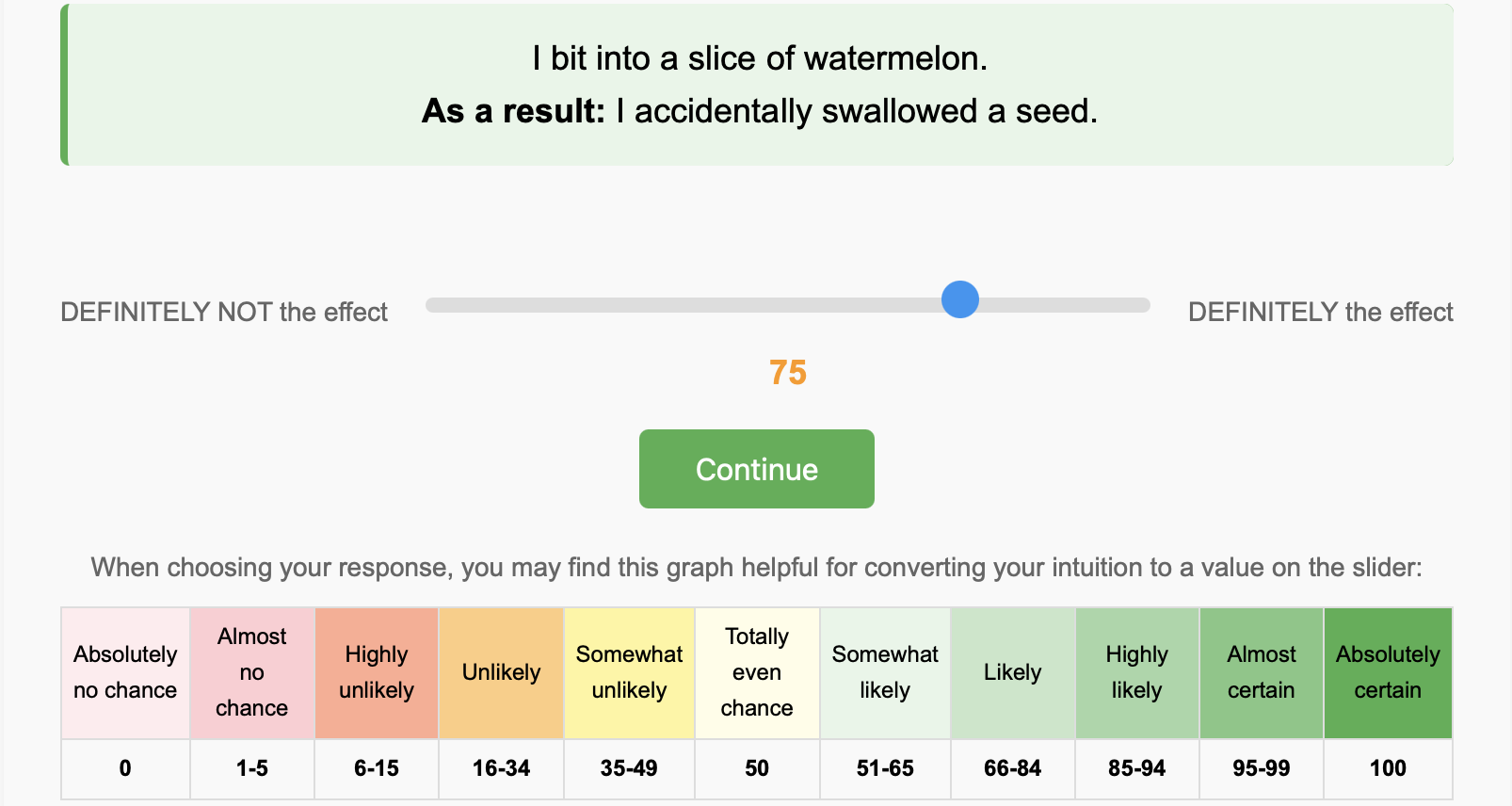}
    \caption{Screenshot of the annotation UI for the main phase of the crowdsourced experiment. The UI and task format follow from what participants were shown in the instructional phase. But at this stage, participants have been informed that unlike in the previous phase, there are no `right' or `wrong' answers.}
    \label{fig:human-annotation-example}
\end{figure*}

Upon completion of this instructional phase, participants were informed that they would now enter the main phase of the experiment, for which there were no `right' or `wrong' answers.
\Cref{fig:human-annotation-example} shows an example of the UI in this main phase. 
Participants were presented with up to 30 \textsc{ProbCOPA} items sequentially (with similarly formatted attention checks interspersed); they provided responses for each, and were given no further feedback.
At the end of the experiment, participants were given the chance to raise any comments or questions about how the experiment was conducted; we received no feedback indicating any difficulty with the task.

\begin{figure*}
    \centering
    \includegraphics[width=\linewidth]{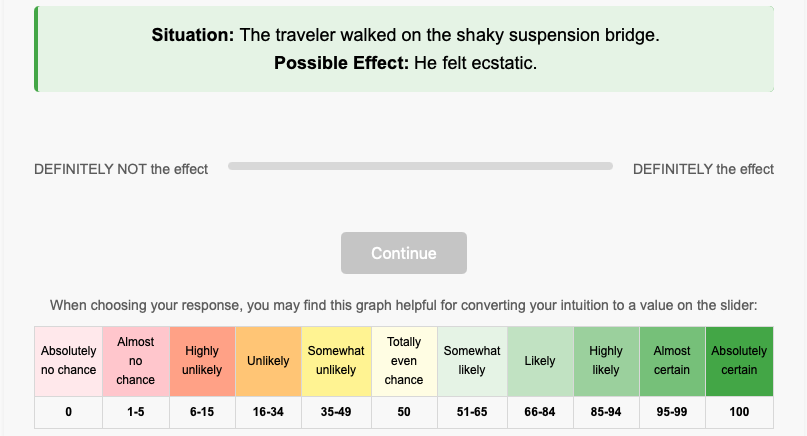}
    \caption{Screenshot of the annotation UI for our validation experiment, in which we slightly vary the prompt wording to closer align with the wording models are presented with (see \Cref{subsec:human-reproducibility}). Note that this variation does not produce different response distributions from our original annotations.}
    \label{fig:human-annotation-example-prompt-variation}
\end{figure*}

As mentioned in \Cref{subsec:human-reproducibility}, we also ran two rounds of human validation after obtaining our original annotations.
In the first, we re-ran the exact same experiment with 30 new participants (on a subset of the data); in the second, we also adjusted the prompt wording slightly.
\Cref{fig:human-annotation-example-prompt-variation} shows an example from this second round of human validation, where we align the exact task wording more closely with the prompt provided to LLMs (see Appendix \ref{appendix:prompt}).
Note that in both rounds of human validation, we obtained response distributions that were not statistically significantly different from our original annotations (see \Cref{subsec:human-reproducibility}). 

\section{Model Inference Details}
\label{appendix:model-details}

\begin{table}[h]
\centering
\scriptsize
\begin{tabular}{p{0.275\columnwidth}p{0.625\columnwidth}}
\toprule
Model Name & Exact Model Version \\
\midrule
Gemini-3 & \texttt{gemini-3-pro-preview} \\
GPT-5 & \texttt{gpt-5-2025-08-07} \\
Claude Sonnet-4.5 & \texttt{claude-sonnet-4-5-20250929} \\
Qwen3 & \texttt{Qwen3-235B-A22B-Thinking-2507} \\
Kimi-K2 & \texttt{Kimi-K2-Thinking} \\
GLM-4.6 & \texttt{GLM-4.6} \\
Grok-4.1 Fast & \texttt{grok-4.1-fast} \\
DeepSeek-R1 & \texttt{DeepSeek-R1} \\
\bottomrule
\end{tabular}
\caption{Exact model versions used in this study.}
\label{tab:model-versions}
\end{table}

\Cref{tab:model-versions} shows the exact models used in this study.
We ran inference on Gemini-3 using the Gemini API; GPT-5 using the OpenAI API; Claude Sonnet-4.5 using the Anthropic API; and Qwen3, Kimi-K2, GLM-4.6 and DeepSeek-R1 using the Together AI API. 
For all of these models, we made API calls using each respective provider's Batch API functionality.
We ran inference on Grok-4.1 Fast using OpenRouter's API (which did not offer a Batch API functionality).

\begin{figure*}
    \centering
    \includegraphics[width=\linewidth]{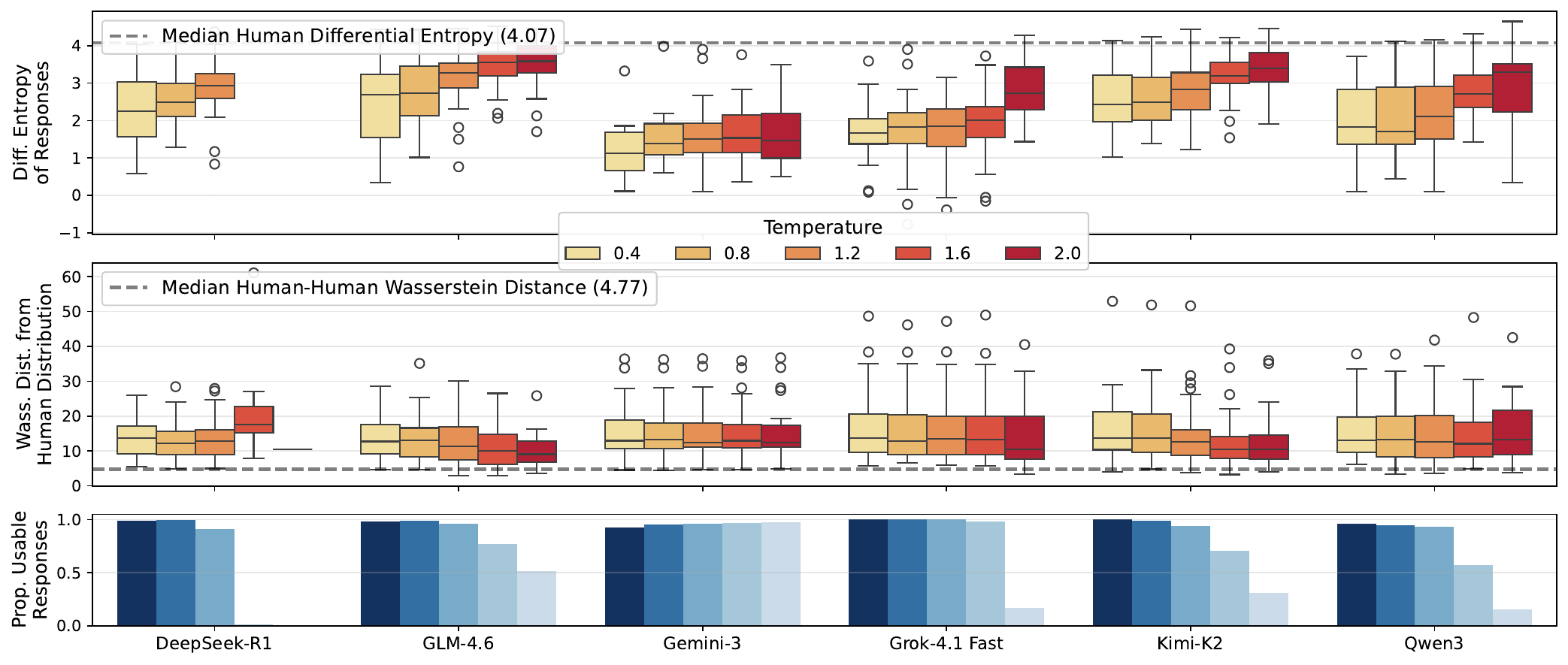}
    \caption{Results from prompting each of the models tested with different temperature settings. \textbf{Top row:} Distributions of the differential entropy of likelihood scores generated by models for each item. \textbf{Middle row:} Distributions of item-level Wasserstein distances between model and human likelihood score distributions. \textbf{Bottom row:} Proportion of responses with a final likelihood score returned within the maximum token limit (4224). Increasing temperature does lead to more diverse responses from models (top row), and for some models, closer alignment with human response distributions (middle row). But this comes at the cost of far fewer responses containing usable responses (bottom row; many responses at higher temperature values devolve into endless sequences of random tokens).}
    \label{fig:unified-temperature-ablation}
\end{figure*}

Temperature values were set to model defaults except when running temperature experiments (see \Cref{fig:unified-temperature-ablation}, \Cref{subsec:model-results-results}).
Reasoning effort was set to `medium' for GPT-5, Qwen3, Kimi-K2, GLM-4.6 and DeepSeek-R1 (which take this argument) while `thinking budget' was set to 1024 for Claude Sonnet-4.5 and Gemini-3 (which take this argument instead)\textemdash once again, except when running reasoning effort experiments (see \Cref{subsec:model-results-results}).
Note that although the Gemini API documentation suggests that the model accepts a `reasoning effort' parameter, at the time of running our experiments, this had not been implemented in the batch API functionality\textemdash which is why we instead controlled the `thinking budget' parameter for the model.
Maximum new token limits were set to 2048 for the main experiment and persona prompting experiments, and increased to 4224 for the temperature and reasoning effort experiments.

\section{Prompt to Models}
\label{appendix:prompt}
Below is the exact prompt we provide to the reasoning LLMs we test in this study.
Note that the scale provided in the system prompt is the same scale provided to humans in the annotation procedure.

\vspace{2em}
\textbf{Prompt:}

\vspace{1em}
\texttt{Consider the following situation and possible effect.}

\vspace{1em}
\texttt{Situation: \{premise\}}

\texttt{Possible Effect: \{hypothesis\}}

\vspace{1em}
\texttt{Given the situation, how likely is this effect?}
\texttt{Respond with a numerical value between 0 and 100, where 0 indicates that this is DEFINITELY NOT the effect, and 100 indicates that this is DEFINITELY the effect.}

\vspace{1em}
\textbf{System Prompt:}

\vspace{1em}
\texttt{You provide responses to questions about the likelihood of an effect given some situation.}
\texttt{After any internal reasoning, reply with a single number between 0 and 100, enclosed in <answer> tags.}
\texttt{You can use the following descriptions of numerical ranges to help guide your response:}

\texttt{0: Absolutely no chance}

\texttt{1-5: Almost no chance}

\texttt{6-15: Highly unlikely}

\texttt{16-34: Unlikely}

\texttt{35-49: Somewhat unlikely}

\texttt{50: Totally even chance}

\texttt{51-65: Somewhat likely}

\texttt{66-84: Likely}

\texttt{85-94: Highly likely}

\texttt{95-99: Almost certain}

\texttt{100: Absolutely certain}

\section{Persona Prompting}
\label{app:persona-prompting}
\begin{figure*}
    \centering
    \includegraphics[width=\linewidth]{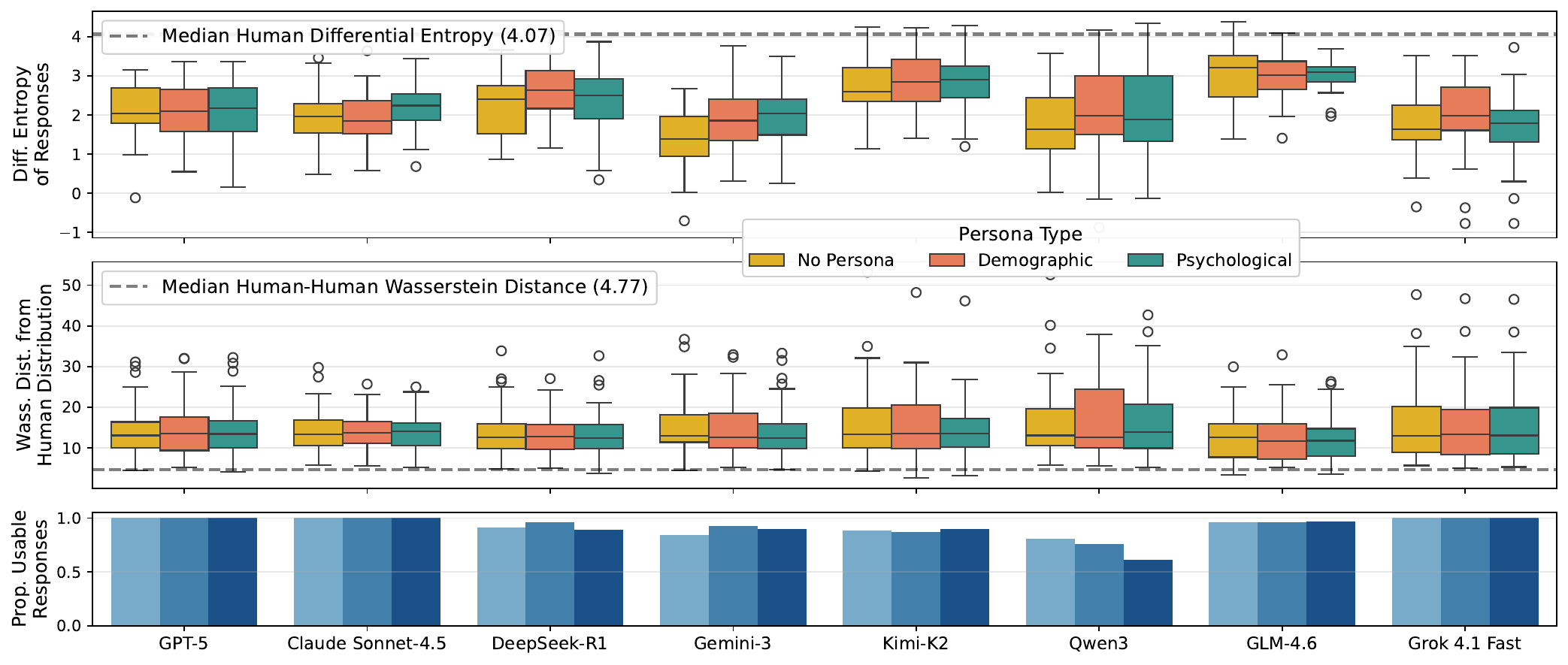}
    \caption{Results from prompting each of the models tested, using a different persona each time a response is sampled for a given \textsc{ProbCOPA} item. \textbf{Top row:} Distributions of the differential entropy of likelihood scores generated by models for each item. \textbf{Middle row:} Distributions of item-level Wasserstein distances between model and human likelihood score distributions. \textbf{Bottom row:} Proportion of responses with a final likelihood score returned within the maximum token limit (2048). Having models adopt different personas\textemdash whether these personas are based on demographic or psychological profiles (see Appendix \ref{app:persona-prompting})\textemdash does sometimes lead to slightly more response variation, but fails to simulate human-level response variation (top row), nor human-like response distributions (middle row).}
    \label{fig:unified-persona-prompting-ablation}
\end{figure*}

As mentioned in \Cref{subsec:model-results-results}, we run a follow-up experiment on a subset of \textsc{ProbCOPA}, in which we prompt the models with different persona descriptions, to test whether this yields more human-like response distributions.
For each of the 30 responses we sample from a model on a single \textsc{ProbCOPA} item (see \Cref{subsec:human-results-methodology}), we append to the system prompt (see Appendix \ref{appendix:prompt}) a different persona description, that specifies either a demographic or psychological description. 
See \Cref{tab:personas} for examples.
As \Cref{fig:unified-persona-prompting-ablation} shows, such persona prompting fails to provide human-level response variation or human-like response distributions.

\section{Claude Opus-4.6}
\label{app:claude-opus-4-6}

We attempted to also test Claude Opus-4.6 \citep{anthropic2026opus46} on \textsc{ProbCOPA}, using Anthropic's Batch API functionality the same way as we did to test Claude Sonnet-4.5. 
Doing so, however, yielded significantly different results. 
Most notably, for each item we tested (the same subset as we used for temperature and reasoning effort experiments), Claude Opus-4.6 returned almost completely invariant likelihood scores across its 30 sampled responses, and almost never with a reasoning chain summary.
We show these findings in \Cref{tab:claude-opus-4-6}.
Although there is somewhat more diversity in responses under the `high' reasoning effort condition, this is nevertheless limited.

Crucially, we find that the number of output tokens (which includes the original reasoning tokens we do not get access to) is always exactly 10 under the `medium' and `low' reasoning effort conditions, and often the same even under the `high' reasoning effort condition.
We speculate that this relates to Claude Opus-4.6 using a so-called `adaptive' thinking budget \citep{anthropic2026opus46}.
It is possible that the model (or some auxiliary system used by the API) classifies most of our inputs as not actually requiring a reasoning chain to solve, and that we are therefore getting direct responses from the model, without any meaningful intermediate reasoning chain.
Without further transparency into the model or the API that is used to access it, however, all of this remains only speculative.
In view of the lack of clarity around how to interpret these results, we exclude them from our main analysis, and instead report them here.

\section{Extended Results Figures}
\label{app:extended-results}

\Cref{fig:pavlick-comparison,fig:model-response-distributions-all,fig:reasoning-chain-length-human-diff-entropy-all,fig:reasoning-chain-length-human-response-time-all,fig:model-human-wasserstein-distance-all,fig:model-human-median-responses-all,fig:model-human-entropy-all,fig:unified-temperature-ablation,fig:unified-reasoning-effort-ablation,fig:unified-persona-prompting-ablation} and \Cref{tab:reasoning-chain-correlations} below show extended results referred to in the main body of this paper.

\clearpage

\begin{table*}[h]
\centering
\footnotesize
\renewcommand{\arraystretch}{1.25}
\begin{tabularx}{0.95\textwidth}{>{\raggedright\arraybackslash}p{0.12\textwidth}
                               >{\raggedright\arraybackslash}p{0.20\textwidth}
                               >{\raggedright\arraybackslash}p{0.18\textwidth}
                               >{\raggedright\arraybackslash}p{0.18\textwidth}
                               >{\raggedright\arraybackslash}p{0.15\textwidth}}
\hline
{Reasoning Effort} & {Proportion of non-Empty Reasoning Summaries} & {Range of Total Count of Output Tokens} & {Mean Differential Entropy of Responses} & {Median Number of Unique Responses} \\
\hline
\texttt{low}    & 0.000 & 10--10     & -0.486 & 1.0 \\
\texttt{medium} & 0.000 & 10--10     & -0.667 & 1.0 \\
\texttt{high}   & 0.198 & 10--250    & -0.296 & 1.0 \\
\hline
\end{tabularx}
\caption{Preliminary results from Claude Opus-4.6. Under `low' and `medium' reasoning effort settings, the model always returns empty reasoning chain summaries, with the total number of output tokens being always exactly 10. Under the `high' reasoning effort setting, we see a small proportion of responses include non-empty reasoning chain summaries, and a wider range of output token counts across all model responses. However, we still see almost zero variability in sampled model responses, with the mean differential entropy of item-wise responses being negative (a mathematical quirk of differential entropy on distributions with near-zero variance). Similarly, on average, only one unique likelihood score is provided across the 30 sampled responses for a given \textsc{ProbCOPA} item.}
\label{tab:claude-opus-4-6}
\end{table*}

\begin{table*}
    \centering\footnotesize
    \begin{tabular}{p{0.15\textwidth} p{0.75\textwidth}}
        \toprule
        \textbf{Persona Type} & \textbf{Persona Prompt Examples} \\
        \midrule
        \textit{Demographic}         & \texttt{You are a 23-year-old female barista in Ottawa, who is saving up to backpack across Europe. Your first language is English.} \\[18pt]
                            & \texttt{You are a 58-year-old male factory worker in Detroit, who is nearing retirement after three decades in the auto industry. Your first language is English.} \\[24pt]
                            & \texttt{You are a 36-year-old male taxi driver in Birmingham, who works night shifts to have more time with his young children during the day. Your first language is English.} \\
        \midrule
        \textit{Psychological}           & \texttt{You balance curiosity with pragmatism, comfortable with both the familiar and the new. You are dependable and organized, often making plans to reach your goals. You feel comfortable in both quiet and social settings, adapting as needed. You are caring and cooperative, often attuned to the feelings of those around you. You experience typical ups and downs but manage emotions with balance.} \\[68pt]
                                & \texttt{You are imaginative and drawn to creativity, often seeking new ways to challenge your thinking. You manage responsibilities reasonably well without being overly rigid. You feel comfortable in both quiet and social settings, adapting as needed. You balance kindness with self-interest, showing empathy without overextending yourself. You sometimes feel self-doubt and worry, but this also makes you sensitive to others’ struggles.} \\
        \bottomrule
    \end{tabular}
    \caption{Examples of persona descriptions used in our persona prompting experiments. \textit{Demographic} persona prompts attempted to simulate some of the demographic variability in our human annotator pool (see \Cref{subsec:human-annotation}). \textit{Psychological} descriptions, on the other hand, attempted to simulate variation in personality. Neither type of persona prompting yielded human-level variation or human-like response distributions (see \Cref{fig:unified-persona-prompting-ablation}).}
    \label{tab:personas}
\end{table*}

\clearpage

\begin{figure*}
    \centering
    \includegraphics[width=\linewidth]{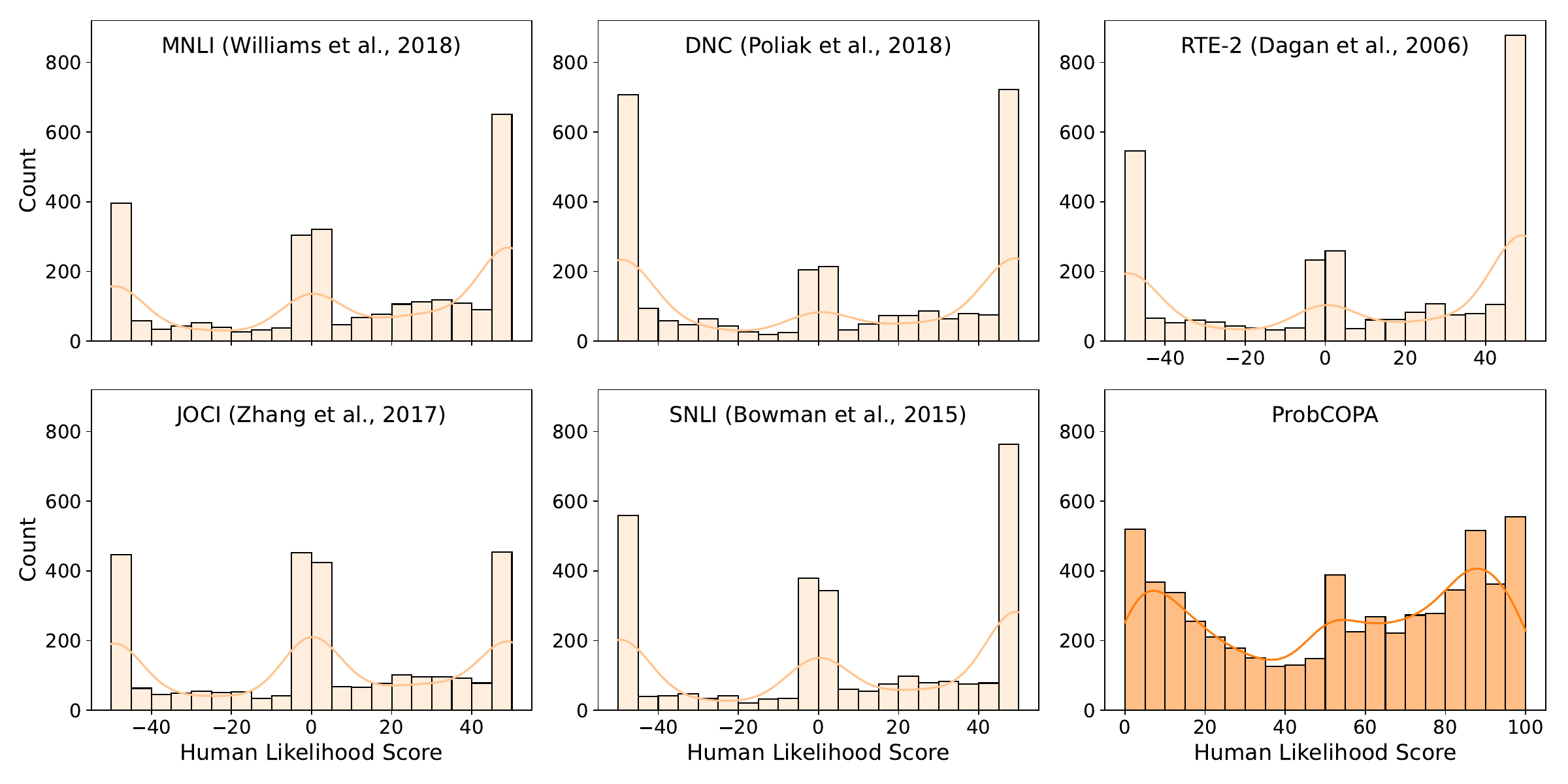}
    \caption{Overall human likelihood score distribution across (i) five major NLI datasets, collected by \citet{pavlick2019inherent}, and (ii) \textsc{ProbCOPA} (ours). Likelihood scores collected by \citet{pavlick2019inherent} for the five major NLI datasets lie on a scale from $-50$ (hypothesis definitely false given premise) to $50$ (hypothesis definitely true given premise). All datasets are subject to tri-modal distributions; but \textsc{ProbCOPA} items receive far more annotations that lie in between these three modes, indicating more graded, probabilistic judgments than for other NLI datasets.}
    \label{fig:pavlick-comparison}
\end{figure*}

\begin{figure*}
    \centering
    \includegraphics[width=\linewidth]{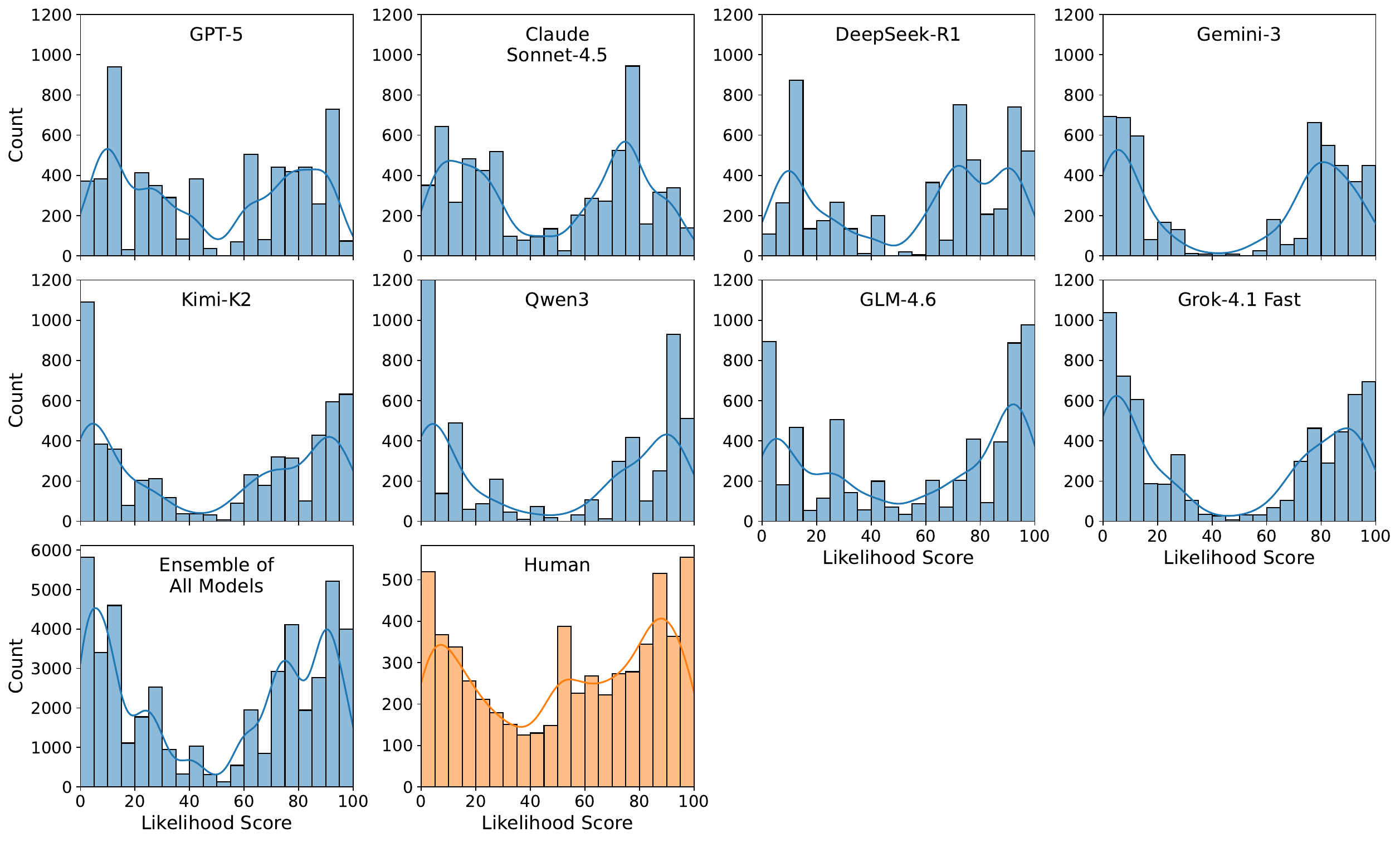}
    \caption{Distribution of likelihood scores across all \textsc{ProbCOPA} items, from all models tested, contrasted against the same distribution from humans. While humans yield an overall likelihood score distribution that is tri-modal, with a large number of responses towards the middle range of the scale, models yield an overall distribution that is bi-modal, with few likelihood scores in the middle range of the scale.}
    \label{fig:model-response-distributions-all}
\end{figure*}

\begin{table*}[htbp]
\centering
\footnotesize
\begin{tabular}{p{0.25\textwidth} p{0.15\textwidth}p{0.15\textwidth} p{0.15\textwidth}p{0.15\textwidth}}
\toprule
& \multicolumn{2}{p{0.3\textwidth}}{Reasoning chain length (tokens) \hspace{2em} $\sim$  Differential entropy of human likelihood scores} & \multicolumn{2}{p{0.3\textwidth}}{Reasoning chain length (tokens) \hspace{1em} $\sim$ Human response time (log-transformed and $z$-scored)} \\
\cmidrule(lr){2-3} \cmidrule(lr){4-5}
Model & Spearman's $\rho$ & $p$-value & Spearman's $\rho$ & $p$-value \\
\midrule
GPT-5 & 0.30 & 7.06e-06 & 0.17 & 1.19e-02 \\
Claude Sonnet-4.5 & 0.18 & 9.61e-03 & 0.12 & 8.74e-02 \\
DeepSeek-R1 & 0.36 & 6.54e-08 & 0.18 & 9.85e-03 \\
Gemini-3 & 0.50 & 2.10e-14 & 0.18 & 9.86e-03 \\
Kimi-K2 & 0.33 & 1.05e-06 & 0.24 & 4.44e-04 \\
Qwen3 & 0.27 & 6.97e-05 & 0.25 & 2.48e-04 \\
GLM-4.6 & 0.14 & 4.18e-02 & -0.02 & 8.22e-01 \\
Grok-4.1 Fast* & NA & NA & NA & NA \\
Ensemble of All Models & 0.44 & 1.90e-11 & 0.23 & 6.35e-04 \\
\bottomrule
\end{tabular}
\caption{Spearman correlations between reasoning chain lengths and (i) the differential entropy of human likelihood scores, and (ii) human response time, log-transformed and by-participant $z$-scored. While correlations between reasoning chain lengths and human likelihood score entropy suggest a relationship between the two for most models, correlations with human response time are consistently lower. *Grok-4.1 Fast does not return reasoning chain information, and is therefore excluded from this analysis; for Claude Sonnet-4.5, we use the number of output tokens as a proxy for the reasoning chain length, since the latter is not directly provided by the API, and the model's final output is only a single token in \texttt{<answer>} tags.}
\label{tab:reasoning-chain-correlations}
\end{table*}

\begin{figure*}
    \centering
    \includegraphics[width=\linewidth]{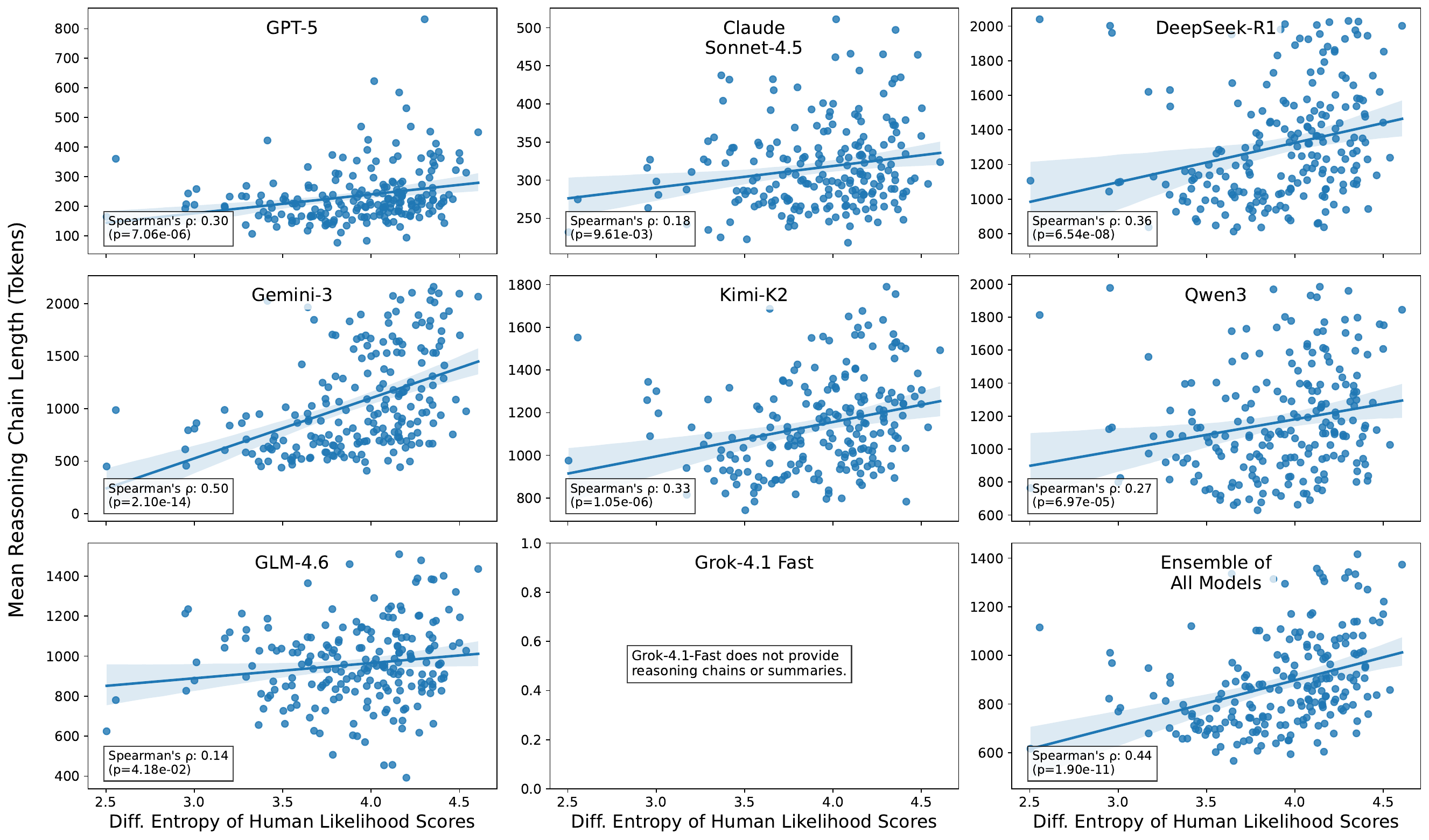}
    \caption{Full set of item-wise comparisons of reasoning chain length and differential entropy of human likelihood scores (correlations shown in \Cref{tab:reasoning-chain-correlations}). Correlations for most models are at least modest ($\rho\geq0.30$), with the highest for Gemini-3. *Grok-4.1 Fast does not return reasoning chain information, and is therefore excluded from this analysis; for Claude Sonnet-4.5, we use the number of output tokens as a proxy for the reasoning chain length, since the latter is not directly provided by the API, and the model's final output is only a single token in \texttt{<answer>} tags.}
    \label{fig:reasoning-chain-length-human-diff-entropy-all}
\end{figure*}

\begin{figure*}
    \centering
    \includegraphics[width=\linewidth]{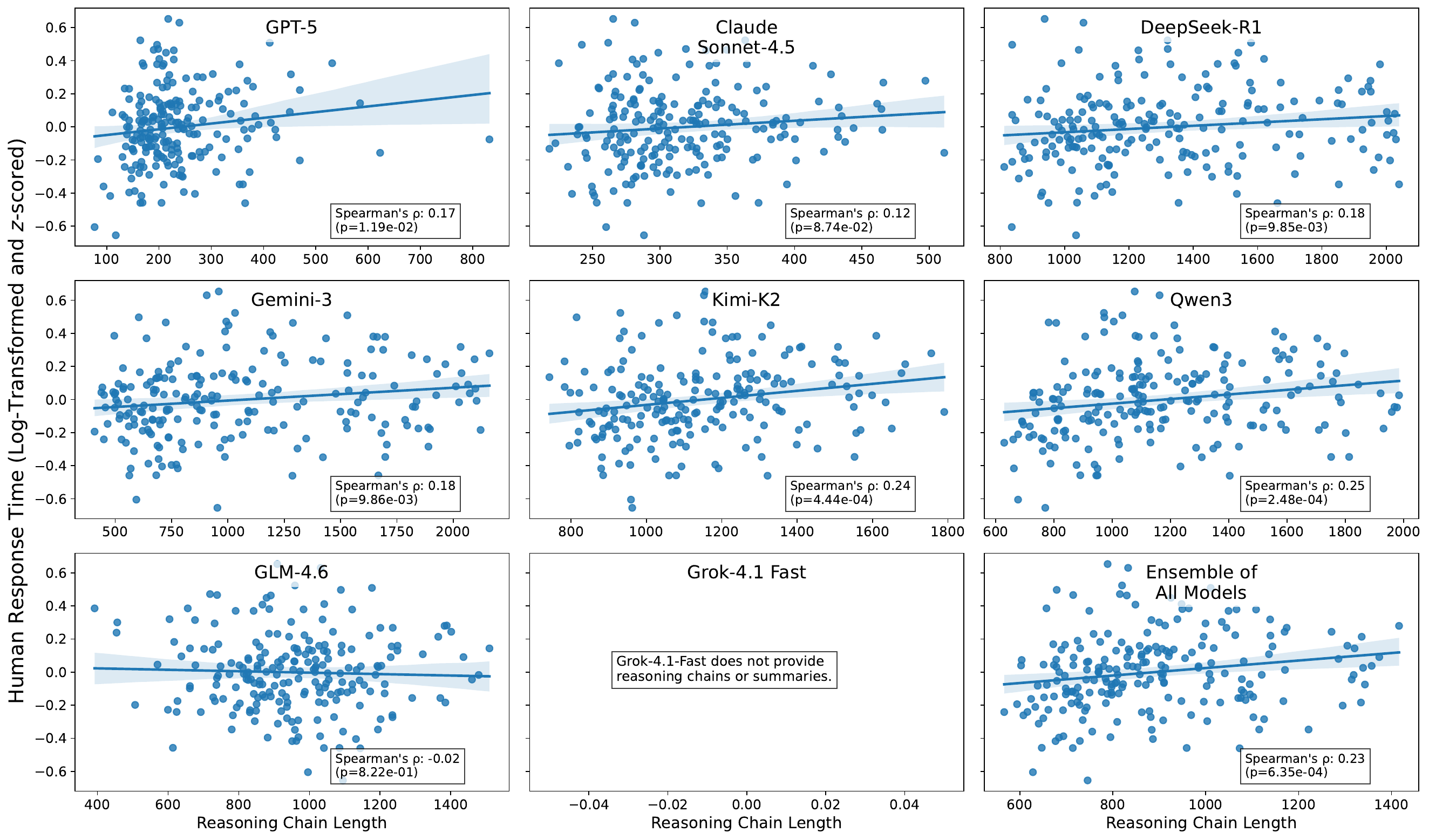}
    \caption{Full set of item-wise comparisons of reasoning chain length and human response time, log-transformed and by-participant $z$-scored (correlations shown in \Cref{tab:reasoning-chain-correlations}). Correlations are consistently lower than when comparing reasoning chain length against human differential entropy. *Grok-4.1 Fast does not return reasoning chain information, and is therefore excluded from this analysis; for Claude Sonnet-4.5, we use the number of output tokens as a proxy for the reasoning chain length, since the latter is not directly provided by the API, and the model's final output is only a single token in \texttt{<answer>} tags.}
    \label{fig:reasoning-chain-length-human-response-time-all}
\end{figure*}

\begin{figure*}
    \centering
    \includegraphics[width=\linewidth]{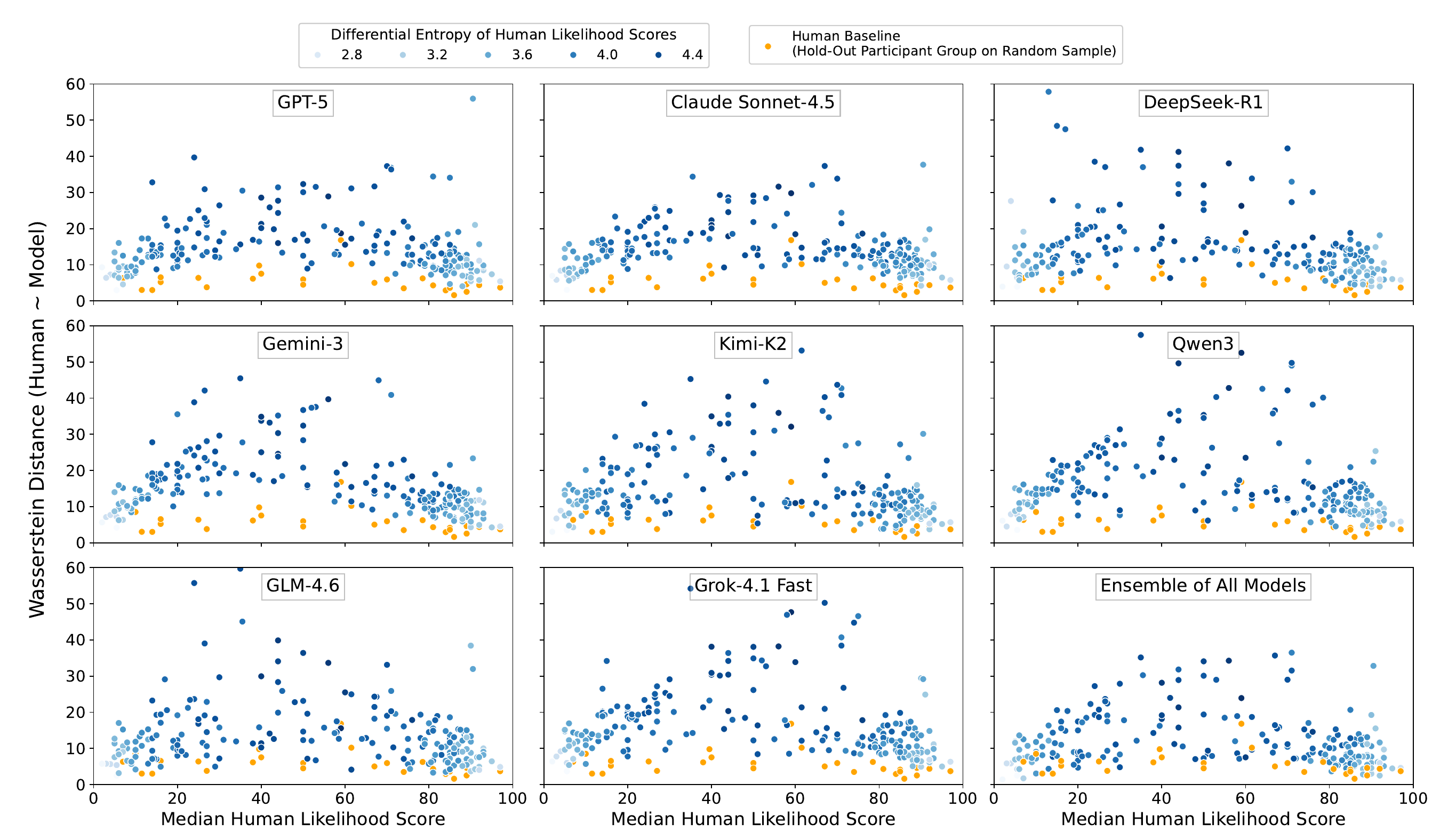}
    \caption{Item-wise Wasserstein distances between likelihood score distributions from original human annotators and each model tested, with the same comparison against human baseline annotations. Wasserstein distances between model and human likelihood scores are highest for items with middle-range median scores from humans (which also have the highest differential entropy of human responses). But no such trade-off exists for human-to-human baseline comparisons, which show consistently higher distributional similarity (shown in lower Wasserstein distances) for almost all items.}
    \label{fig:model-human-wasserstein-distance-all}
\end{figure*}

\begin{figure*}
    \centering
    \includegraphics[width=\linewidth]{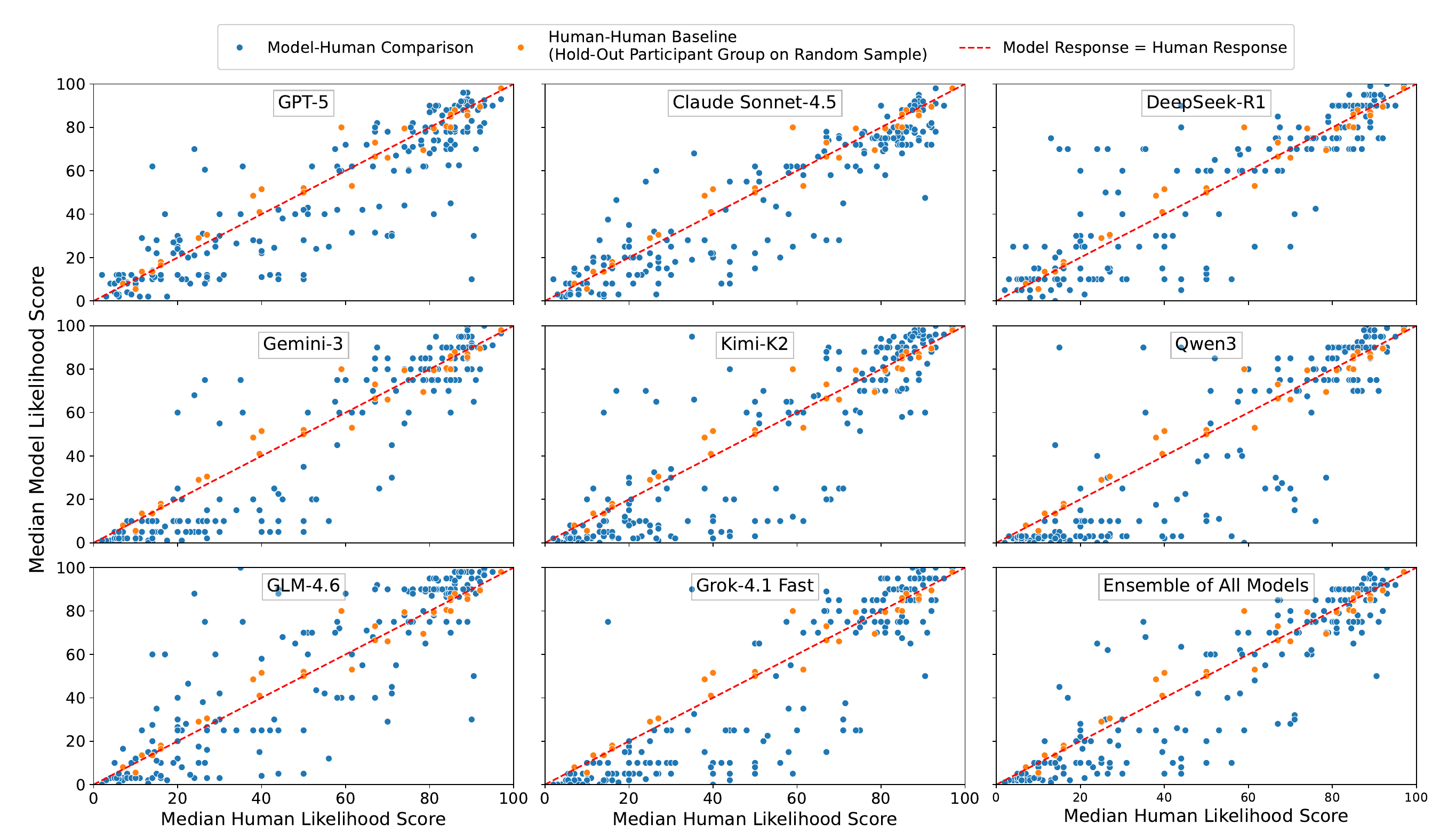}
    \caption{Item-wise median likelihood scores from original annotations and each of the models tested, along with the same comparison against human baseline annotations. While median likelihood scores from humans and models show some similarity at the two extreme ends of the likelihood scale, this relationship breaks down towards the middle\textemdash unlike median scores from our human baseline, which correlate closely with original annotations throughout.}
    \label{fig:model-human-median-responses-all}
\end{figure*}

\begin{figure*}
    \centering
    \includegraphics[width=\linewidth]{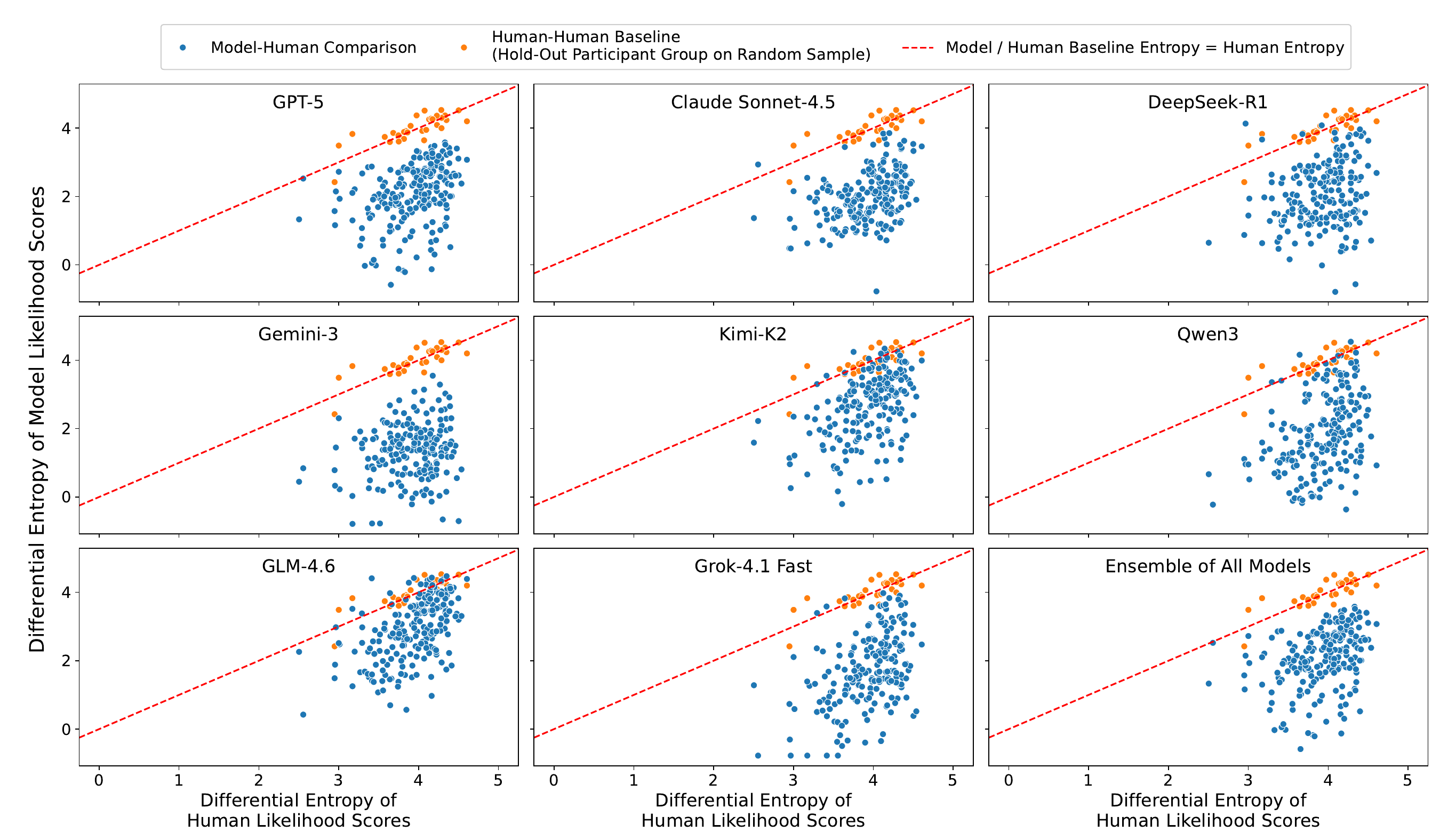}
    \caption{Item-wise differential entropy of likelihood scores from original \textsc{ProbCOPA} annotations and each model tested, along with the same comparison against human baseline annotations. While differential entropy from our human baseline is roughly similar to those from the original annotations, item-level differential entropy of likelihood scores is almost always higher for humans than models.}
    \label{fig:model-human-entropy-all}
\end{figure*}

\begin{figure*}
    \centering
    \includegraphics[width=\linewidth]{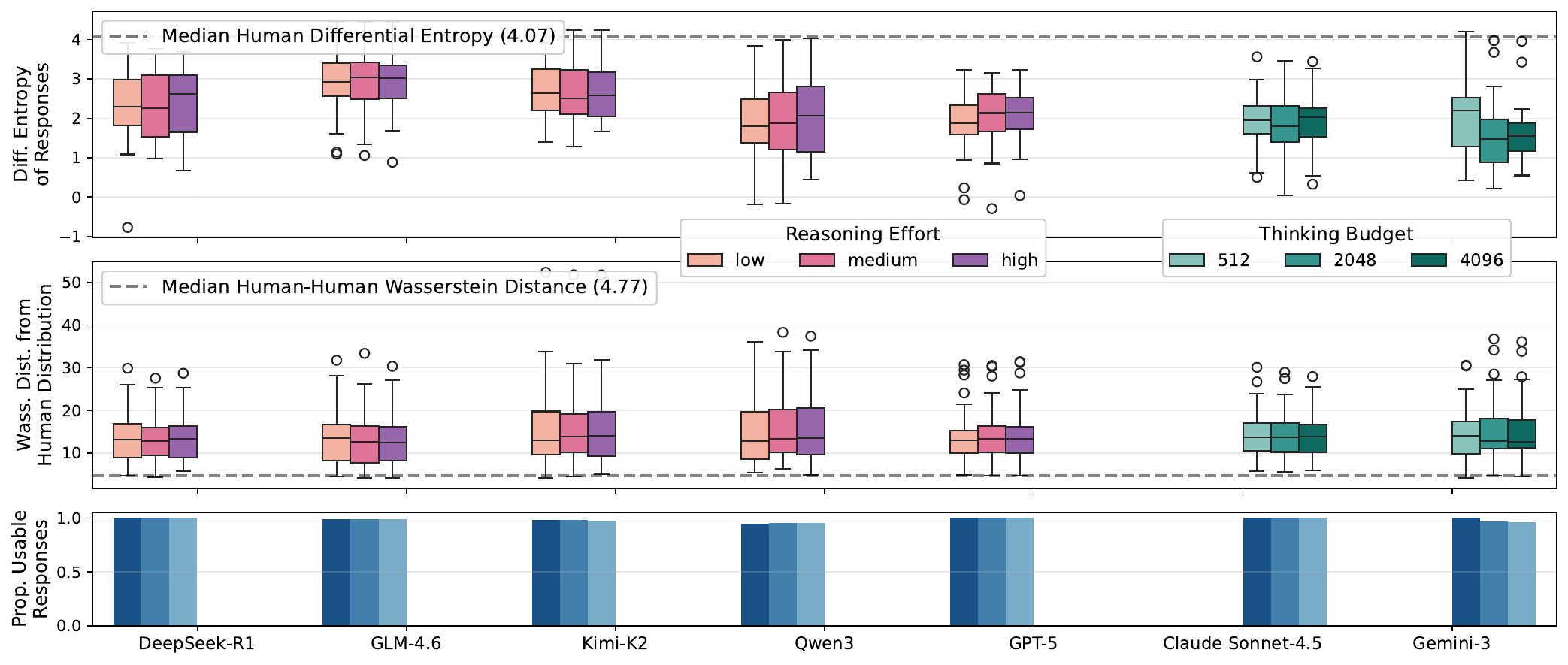}
    \caption{Results from prompting each of the models tested with different reasoning effort / `thinking budget' settings. \textbf{Top row:} Distributions of the differential entropy of likelihood scores generated by models for each item. \textbf{Middle row:} Distributions of item-level Wasserstein distances between model and human likelihood score distributions. \textbf{Bottom row:} Proportion of responses with a final likelihood score returned within the maximum token limit (4224). Increasing reasoning effort does not appear to lead to any meaningful differences in model likelihood score distributions, as confirmed in \Cref{subsec:model-results-results}.}
    \label{fig:unified-reasoning-effort-ablation}
\end{figure*}

\clearpage

%%%%%%%%%%%%%%%%%%%%%%%%%%%%%%%%%%%%%%%%%%%%%%%%%%
%%%%%%%%%% <HUMAN INSTRUCTIONS SCREENSHOT> %%%%%%%%
%%%%%%%%%%%%%%%%%%%%%%%%%%%%%%%%%%%%%%%%%%%%%%%%%%
%%%%%%%%%%%%%%%%%%%%%%%%%%%%%%%%%%%%%%%%%%%%%%%%%%
%%%%%%%%%% /<HUMAN INSTRUCTIONS SCREENSHOT> %%%%%%%%
%%%%%%%%%%%%%%%%%%%%%%%%%%%%%%%%%%%%%%%%%%%%%%%%%%

%%%%%%%%%%%%%%%%%%%%%%%%%%%%%%%%%%%%%%%%%%%%%%%%%%
%%%%%%%%%% <HUMAN INSTRUCTIONS SCREENSHOT (WRONG)> %%%%%%%%
%%%%%%%%%%%%%%%%%%%%%%%%%%%%%%%%%%%%%%%%%%%%%%%%%%
%%%%%%%%%%%%%%%%%%%%%%%%%%%%%%%%%%%%%%%%%%%%%%%%%%
%%%%%%%%%% /<HUMAN INSTRUCTIONS SCREENSHOT (WRONG)> %%%%%%%%
%%%%%%%%%%%%%%%%%%%%%%%%%%%%%%%%%%%%%%%%%%%%%%%%%%

%%%%%%%%%%%%%%%%%%%%%%%%%%%%%%%%%%%%%%%%%%%%%%%%%%
%%%%%%%%%% <HUMAN ANNOTATION SCREENSHOT> %%%%%%%%
%%%%%%%%%%%%%%%%%%%%%%%%%%%%%%%%%%%%%%%%%%%%%%%%%%
%%%%%%%%%%%%%%%%%%%%%%%%%%%%%%%%%%%%%%%%%%%%%%%%%%
%%%%%%%%%% /<HUMAN ANNOTATION SCREENSHOT> %%%%%%%%
%%%%%%%%%%%%%%%%%%%%%%%%%%%%%%%%%%%%%%%%%%%%%%%%%%

%%%%%%%%%%%%%%%%%%%%%%%%%%%%%%%%%%%%%%%%%%%%%%%%%%
%%%%%%%%%% <HUMAN ANNOTATION SCREENSHOT> %%%%%%%%
%%%%%%%%%%%%%%%%%%%%%%%%%%%%%%%%%%%%%%%%%%%%%%%%%%
%%%%%%%%%%%%%%%%%%%%%%%%%%%%%%%%%%%%%%%%%%%%%%%%%%
%%%%%%%%%% /<HUMAN ANNOTATION SCREENSHOT> %%%%%%%%
%%%%%%%%%%%%%%%%%%%%%%%%%%%%%%%%%%%%%%%%%%%%%%%%%%

\end{document}